\documentclass{article} %
\usepackage{iclr2025_conference_arxiv,times}

\usepackage{amsmath,amsfonts,bm}

\def\eqref#1{equation~\ref{#1}}

\def\1{\bm{1}}

\DeclareMathAlphabet{\mathsfit}{\encodingdefault}{\sfdefault}{m}{sl}
\SetMathAlphabet{\mathsfit}{bold}{\encodingdefault}{\sfdefault}{bx}{n}

\usepackage{hyperref}
\usepackage{url}
\usepackage{graphicx}
\usepackage{algorithm} 
\usepackage{algpseudocode} 
\usepackage{enumitem}
\usepackage{booktabs}
\usepackage{multirow}
\usepackage[frozencache,cachedir=.]{minted}
\usepackage{tcolorbox}
\usepackage{subfigure}
\usepackage{listings}
\usepackage{xspace}

\lstdefinestyle{mystyle}{
    basicstyle=\ttfamily\tiny,
    breakatwhitespace=true,
    breaklines=true,
    keepspaces=true,
    showspaces=false,
    showstringspaces=false,
    frame=single,
    extendedchars=false,
    inputencoding=utf8,
    captionpos=b
}
\title{Self-Explained Keywords Empower Large Language Models for Code Generation}

\author{Lishui Fan, Mouxiang Chen, Zhongxin Liu\thanks{Corresponding author.}  \\
The State Key Laboratory of Blockchain and Data Security\\
Zhejiang University\\
\texttt{\{flscode,chenmx,liu\_zx\}@zju.edu.cn} \\
}

\newcommand{\SEK}{SEK\xspace}

\iclrfinalcopy %
\begin{document}

\maketitle

\begin{abstract}
Large language models (LLMs) have achieved impressive performance in code generation. 
However, due to the long-tail distribution of LLMs' training data, low-frequency terms are typically underrepresented in the training process. Consequently, LLMs often misunderstand or overlook problem-specific, low-frequency keywords during code generation, compromising the accuracy of the generated code.
To address this, we propose a novel technique named \SEK(\textbf{S}elf-\textbf{E}xplained \textbf{K}eywords), which empowers an LLM for better code generation by extracting and explaining the key terms in the problem description with the LLM itself and ranking them based on frequency.
Comprehensive experiments across three benchmarks, i.e., HumanEval(+), MBPP(+), and APPS, with five representative LLMs, show that \SEK can significantly improve LLMs in code generation, yielding substantial and consistent gains. For instance, \SEK improves the Pass@1 of DeepSeek-Coder-V2-Instruct from 85.4\% to 93.3\% on the Humaneval benchmark. Further analysis confirms that \SEK enables the LLMs to shift their attention from low-frequency keywords to their corresponding high-frequency counterparts.
\end{abstract}
\section{Introduction}
Code generation aims to generate a code snippet that meets the intent described in natural language. This process can potentially reduce the costs of software development~\citep{xu2022ide,yin2017syntactic,vaithilingam2022expectation}. 
Recently, the notable success of LLMs such as ChatGPT~\citep{OpenAI2022} and Llama-3~\citep{llama3modelcard} has substantially enhanced the state-of-the-art in code generation. These LLMs demonstrate remarkable proficiency in comprehending natural language descriptions and translating them into code snippets.

Despite the remarkable success, their training data suffer from the long-tail distribution problem~\citep{chen2024large,li2024dawn}, which results in insufficient training of the correspondence between some low-frequency terms and their respective code implementations. Consequently, when performing real-world coding tasks, LLMs often struggle to translate certain low-frequency keywords into appropriate code implementations accurately, and may even overlook such keywords, potentially compromising the accuracy of the generated code.

\begin{figure}[t]
\centering
\begin{minipage}{0.48\linewidth}
    \centering
    \resizebox{\linewidth}{!}{
        \includegraphics{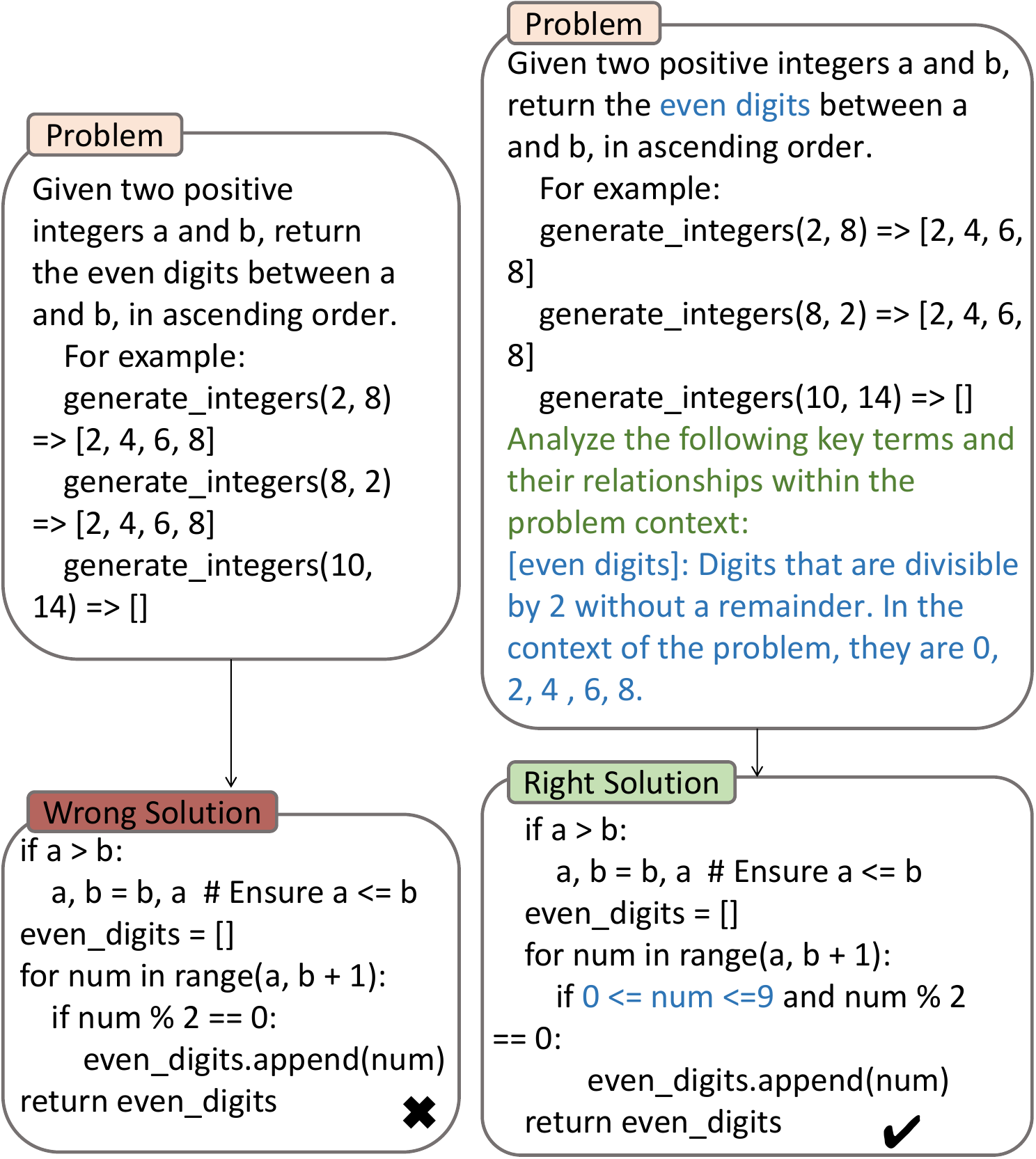}
    }
    \caption{Motivating example.}
    \label{fig_example}
\end{minipage}%
\hspace{0.03\linewidth}%
\begin{minipage}{0.48\linewidth}
    \centering
    \resizebox{\linewidth}{!}{
        \includegraphics{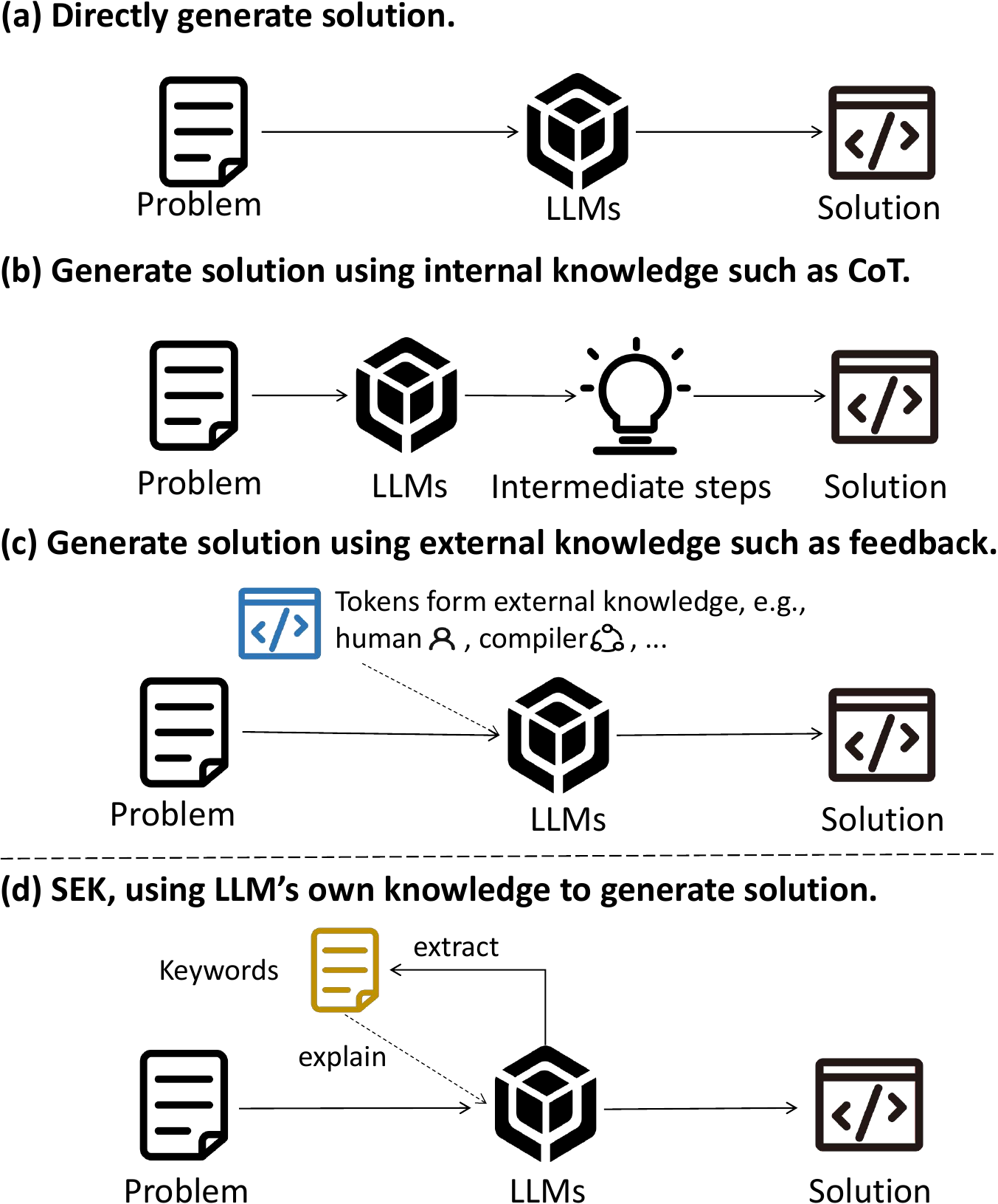}
    }
    \caption{Schematic illustration of various generation approaches.}
    \label{fig_compare}
\end{minipage}
\end{figure}

As illustrated in Figure~\ref{fig_example}, the coding problem requires returning \textit{even digits} within a given range in ascending order. The term \textit{even digits} rarely appears in code datasets, causing LLMs to misinterpret it as \textit{even numbers}. As a result, LLMs fail to recognize that it refers to the even numbers within the range of 0 to 9. However, if we explicitly explain \textit{even digits} using common, high-frequency terms and prompt the LLM to focus on it, the LLM can successfully produce a correct implementation.

However, converting low-frequency keywords into high-frequency natural language descriptions requires additional human effort for identification and explanation. Our key idea is that although the direct mapping from low-frequency terms to code is rare in \textit{code} corpora, the semantics of these terms are typically understandable by LLMs after pre-training on large-scale \textit{general} corpora. Moreover, previous works find that such problem-specific key items can be identified by LLMs~\citep{fang2024llm,fan2024prompt}. Based on this, this work proposes \textbf{S}elf-\textbf{E}xplained \textbf{K}eyword (\SEK), a novel technique leveraging LLMs' own comprehension capabilities to automatically identify and explain low-frequency, problem-specific terms to enhance their understanding of coding problems.

A major challenge in implementing \SEK is how to effectively identify and explain keywords that LLMs might overlook in problem descriptions. To address this, \SEK employs a carefully designed prompt with a few examples, directing LLMs to focus on crucial keywords in the problem description. After extracting the keyword set, the next challenge is how to enhance the original problem prompt. Since keywords with lower frequency are more likely to be misunderstood or overlooked by LLMs, we use a frequency-based ranking algorithm to sort these keywords, which are then appended to the original problem description to construct an augmented prompt. Overall, this approach aligns with the working process of pragmatic human developers, which use auxiliary tools like blackboards to highlight, explain, and rank important parts of requirements~\citep{andrewthe00}.

By overcoming these challenges, \SEK enhances LLMs' problem-solving capabilities in a novel way, distinguishing itself from previous methods in prompt engineering for code generation. As shown in Figure \ref{fig_compare}, unlike previous approaches that often rely on introducing external knowledge, such as human feedback~\citep{chen2023improving,wu2024fine,dubois2024alpacafarm} or the execution results of LLM-generated solutions~\citep{zhong2024ldb,chen2023teaching,zhong2024debug}, into the input, \SEK operates by distilling additional content from the problem description using the LLM itself. 
Chain of Thought (CoT)~\citep{wei2022chain}, which also utilizes LLMs' inherent knowledge for problem-solving, bears the closest resemblance to \SEK. However, the fundamental strategies of CoT and \SEK are different: CoT guides the LLM to think in a chain-like manner, while \SEK directs the LLM to understand and prioritize key concepts.

We evaluate \SEK with five representative LLMs, including three open-source models and two closed-source models, on three widely used code generation benchmarks. Experimental results demonstrate that \SEK effectively enhances code generation performance. For example, \SEK enables Llama-3.1 to achieve a relative improvement of 8.8\% averaged on the used benchmarks.
Notably, DeepSeek-Coder-V2-Instruct with \SEK significantly outperforms it with standard prompting, achieving state-of-the-art performance on several benchmarks (e.g., HumanEval: 85.4\% to 93.3\%). Furthermore, our ablation studies indicate that the carefully designed prompt and the ranking component of \SEK are effective.
Additionally, our attention analysis reveals that \SEK helps LLMs comprehend low-frequency keywords by redirecting attention to their high-frequency counterparts. Comparative case studies with other baselines further illustrate \SEK's efficacy in enhancing LLMs' understanding of low-frequency, problem-specific keywords.
Our code is in the Supplementary Materials and will be made public after review.

\section{Methodology}
Code generation aims to generate a solution program based on a problem description.
Typically, a problem description includes implementation requirements, and several test cases to help further understand the problem.

\begin{figure}[t]
\centering
\includegraphics[width=\textwidth]{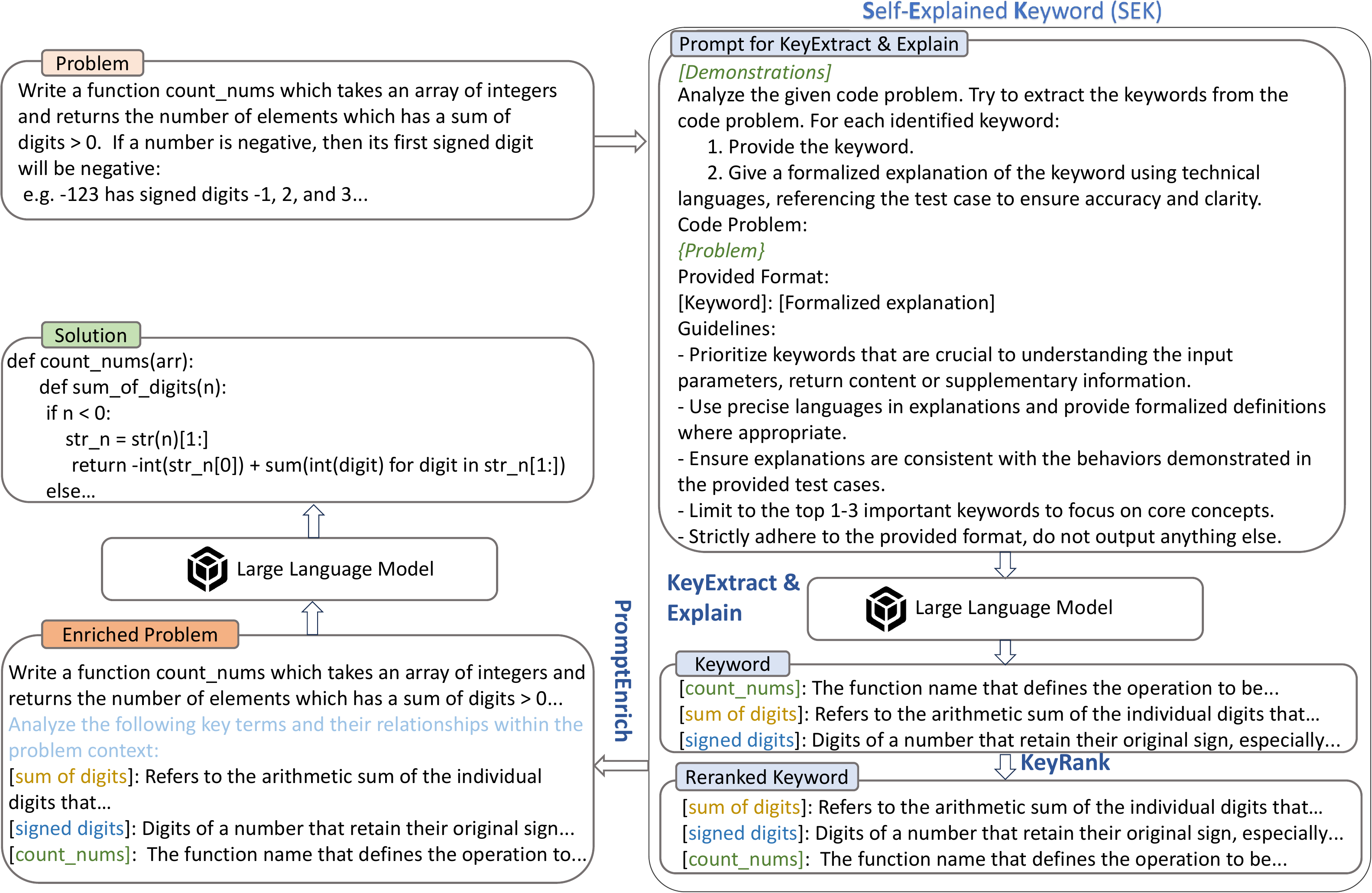}
\caption{The overview of Self-Explained Keyword. The details in each step are omitted.}
\label{fig_overall}
\end{figure}

Figure \ref{fig_overall} illustrates the overview of \SEK. \SEK is designed to address the issue of LLMs overlooking low-frequency terms in the program description due to the long-tail distribution in their training data. To address it, one key is to leverage the LLM's capabilities to identify and explain potentially overlooked keywords within the problem description. We employ a carefully crafted prompt with a few-shot learning method to achieve this.
After obtaining the keywords and their explanations, another challenge is how to effectively integrate them with the original problem description. For this purpose, we introduce a frequency-based ranking algorithm that prioritizes less frequent tokens, which are more likely to be overlooked by the LLM. These ranked keywords are then appended to the original problem description, serving to guide the LLM towards generating an accurate solution.
The process comprises three main steps:

\textbf{KeyExtract \& Explain} (Section \ref{sec_keyextarct}): Based on the problem description, \SEK constructs a prompt to guide the LLM to identify and explain keywords within the problem description.

\textbf{KeyRank} (Section \ref{sec_keyrank}):  \SEK employs a frequency-based ranking algorithm to prioritize the extracted keywords.

\textbf{PromptEnrich} (Section \ref{sec_keysolve}), \SEK concatenates the ranked keywords and their explanations with the original problem description to create an enriched problem description. This comprehensive formulation serves as the final input for the LLM to generate code solutions.

\subsection{KeyExtract \& Explain}
\label{sec_keyextarct}
In this step, \SEK extracts and explains keywords from the given problem description. Our key insight is that LLMs inherently possess strong understanding and reasoning abilities after training on large-scale general corpora, enabling them to explain crucial concepts within a problem description.  The effectiveness of using LLMs for keyword extraction has also been demonstrated by recent studies~\citep{maragheh2023llm,lee2023toward}.
Inspired by this insight, \SEK uses the LLM itself to perform the task with a prompt-based approach. 

Specifically, \SEK begins by designing a prompt to instruct an LLM for keyword extraction and explanation. The prompt is shown in \textit{Prompt for KeyExtract \& Explain} in Figure \ref{fig_overall}, which consists of three parts. First, it provides the overall instruction for the task, namely the generation of keywords and their corresponding explanations. Then, it specifies the format of input and output. Finally, it provides detailed guidelines. Intuitively, terms associated with input, output, and supplementary content (i.e., clarifications of keywords or specifications of value ranges) within the problem description are relatively important, as they contain the problem's core elements, objectives, and constraints (Guideline 1). For explanations, given the potential ambiguity in natural language expressions and the clarity of the public test cases, the generated explanations should be both precise and consistent with these test cases (Guideline 2,3).
We also impose limitations on the keyword quantity to guarantee that the LLM identifies and outputs only the important keywords in the problem description (Guideline 4). Ultimately, to facilitate subsequent processing, we further emphasize the output format (Guideline 5). 
Additionally, we use several examples to leverage LLMs' in-context learning ability for understanding and solving this task. 
\subsection{KeyRank}
\label{sec_keyrank}

After extracting and explaining the keywords, the next goal is to enhance the original prompt. Previous research has demonstrated that LLMs are sensitive to the order of tokens in the prompt, known as position bias~\citep{li2024long,yu2024mitigate}. It highlights the need to carefully arrange the extracted keywords. Notably, pragmatic human developers tend to place more important keywords at the beginning in practice~\citep{andrewthe00}. This preference may be reflected in the training dataset, leading LLMs to also focus more on the keywords written at the front. Therefore, we propose a set of heuristic rules to rank keywords by importance, namely \textbf{KeyRank}. The specific Algorithm is provided in the Appendix~\ref{appendix:alg_keyrank}.

We first examine the keywords extracted by two LLMs (Llama 3.1 and DeepSeekCoder-V2) for part of the coding problems in the APPS training set. These keywords can generally be categorized into three types: (1) \textit{Function keywords}, which match the desired function names, such as \texttt{count\_nums} in Figure~\ref{fig_overall}. (2) \textit{General keywords}, which appear in the problem description, like \texttt{sum of digits} in Figure~\ref{fig_overall}. (3) \textit{Abstract keywords}, which do not appear in any input; instead, they are abstract terms summarized from multiple concepts. For example, for two different concepts ``substring before the dot'' and ``substring after the dot'' in the problem description, LLM may combine them into a single keyword \texttt{substring before/after the dot}. The proportions of these three categories are 22.5\%, 59.9\%, and 17.7\%. 

We hypothesize that abstract keywords are the most important, as they encompass explanations across multiple concepts. General keywords refer to single concepts and are of secondary importance, while function Keywords, whose explanations have already appeared in the problem description, are the least important. Therefore, we propose ordering the keywords as \textit{abstract $\rightarrow$ general $\rightarrow$ function}. Appendix~\ref{append:orders} demonstrates that this heuristic combination order yields the best results.

Moreover, since general keywords represent the majority (59.9\%) and LLMs could extract multiple general keywords for a single problem, we further perform an internal ranking of these general keywords. We argue that a keyword is more important if it appears more frequently in the problem description (i.e., higher term frequency). Conversely, if a keyword appears less frequently in a corpus (i.e., lower document frequency), the corresponding code conversion could be more challenging as we stated in the Introduction section, and thus its explanation is more significant. Therefore, we use the TF-IDF, a widely used metric that combines term frequency (TF) and inverse document frequency (IDF), to assess the importance of general keywords. TF-IDF is calculated as follows:

\begin{equation*}
\text{TF-IDF}=\frac{n_{i}}{ {\textstyle \sum_{k}^{}n_k}}\times \log\frac{|D|}{1+|\{j:t_i\in d_j\}|}.  
\end{equation*}

The first term represents TF, where $n_i$ denotes the number of times the keyword appears in the problem description, and the denominator represents the total occurrences of all items with the same number of grams. The second term represents IDF, where $|D|$ is the total number of documents in the corpus, and the denominator represents the number of documents containing the keyword $t_i$.

We adopt the Python subset of the eval-codealpaca-v1~\citep{luo2023wizardcoder} as the corpus for computing document frequency, which is generated by ChatGPT and can partially reflect the distribution of LLMs' training data. In addition, we demonstrate that \SEK is robust across various corpora.

\subsection{PromptEnrich}
\label{sec_keysolve}
After obtaining the ranked keywords and their explanations, \SEK integrates them with the original problem. As shown in the enriched problem in Figure \ref{fig_overall}, \SEK appends the ranking results to the end of the problem, providing additional explanations for key concepts in the problem. It's worth noting that, to maintain the coherence of the problem context, we insert the phrase \textit{``Analyze the following key terms and their relationships within the problem context:''} after the problem. This acts as a semantic buffer, smoothly transitioning from the original problem description to the appended keywords. The enriched problem is then input into the LLM to generate the final solution.

\section{Experimental Setup}

We conduct a series of experiments to evaluate the effectiveness of the proposed approach \SEK. In this section, we describe our experimental setup, including the selected models, benchmarks, evaluation metrics, baselines, and implementation details.

\subsection{Studied LLMs}
We select five representative LLMs to evaluate \SEK, balancing between open-source and proprietary models, as well as covering a range of model sizes and architectures. The open-source models include Llama-3.1-70B-instruct~\citep{dubey2024llama3herdmodels}, which is a dense decoder-only model with 70-billion parameters, Mixtral-8×22B-instruct-v0.1~\citep{jiang2024mixtralexperts}, which is a sparse Mixture-of-Experts (MOE) model having 141-billion total parameters with 39B active, and DeepSeek-Coder-V2-236B-Instruct-0724~\citep{zhu2024deepseek}, which is a sparse MOE model having 236B parameters with 21B active. We access DeepSeek-Coder via DeepSeek-AI's API. For proprietary models, we include GPT-3.5-turbo-0125~\citep{OpenAI2022} and GPT-4o-mini~\citep{OpenAI2024GPT4o}, accessed via OpenAI's API. Detailed specifications for each model are provided in the Appendix \ref{appendix:LLMs}.

\subsection{Benchmarks and Evaluation Metric}

Following previous work~\citep{chencodet,dongself,zhong2024ldb,jiang2023self}, We conduct experiments on three public code generation benchmarks HumanEval(+)~\citep{chen2021evaluating,liu2024your},  MBPP(+)~\citep{austin2021program,liu2024your}, and APPS~\citep{hendrycks2measuring}. Considering the high cost of evaluating the entire APPS test problems and following prior work~\citep{olausson2023self,huang2024knowledge,lecodechain,yang2023zero}, we randomly select 300 problems from the APPS test set for evaluation\footnote{There are three different difficulty levels of problems in APPS, i.e., introductory, interview, and competition.
Specifically, based on the frequency distribution of problems with different difficulty levels, we sample 60, 180, and 60 problems at the introductory, interview and competition levels, respectively.
All tasks are listed in Appendix~\ref{append:random_apps}.}. To mitigate the uncertainty introduced by random sampling, we conduct multiple experiments with different sample seeds. More details are in Appendix~\ref{append:random_apps}. For detailed descriptions of each benchmark, please refer to Appendix~\ref{appendix:benchmark}. We evaluate model performance using the Pass@1 metric, which measures the ability to generate correct solutions in a single attempt. This also aligns with real-world scenarios where developers aim to produce accurate code on the first try.

\subsection{Baselines}

\begin{itemize}[leftmargin=*,noitemsep]
\item[$\bullet$] \textbf{Default LLM}: 
This approach is based on the EvalPlus framework~\citep{liu2024your}, using problems from the benchmark as input to prompt LLMs for code generation.

\item[$\bullet$] \textbf{CoT (Chain-of-Thought)}~\citep{wei2022chain}: 
This approach generates a series of reasoning steps during the solution-generation process for each problem. To ensure comparative fairness, both the CoT baseline and \SEK employ an equal number of demonstrations.
\item[$\bullet$] \textbf{SelfEvolve}~\citep{jiang2023selfevolve}: This approach first uses LLMs to generate problem-specific knowledge and produce initial code solutions based on such knowledge. Then, it iteratively refines code solutions with LLMs based on execution feedback. Notably, SelfEvolve uses different prompt templates for different benchmarks to extract knowledge. Since these prompt templates have been open-sourced, we consistently apply its two-stage prompts on HumanEval (see Appendix \ref{appendix:prompt_evolve}) in our replication process. For a fair comparison, we remove the self-refinement module, and employ the same number of demonstrations as \SEK.
\item [$\bullet$] \textbf{Beam Search}~\citep{wiseman2016sequence}: This approach employs distinct search beams and optimizes selection during the decoding process. Given that \SEK requires two LLM invocations per coding problem (one for generating keywords and explanations, and one for generating the code), we demonstrate the benefit of the process of \SEK by comparing it with performing two searches within the LLM's searching space, i.e., beam search with a beam size of 2.
\end{itemize}

\subsection{Implementation Details}

\textbf{Prompt Design.} It's worth noting that the implementations of \SEK, CoT, and Beam Search are based on the EvalPlus framework. 
Specifically, the only difference between \SEK and Default is the addition of keywords and explanations to the problem description. APPS contains problems in two formats: call-based formatormat. Following previous work~\citep{olausson2023self,inala2022fault,chencodet}, we employ a two-shot prompt to guide the LLM to generate appropriate solutions for different formats. 

\textbf{Demonstration Selection Strategy.} Inspired by previous work~\citep{wei2022chain,mu2023clarifygpt,wang2022self}, we adopt a differentiated strategy that varies based on benchmark complexity (See Appendix \ref{appendix:details}). To reduce bias, we employ an LLM separate from our target LLMs (Claude-3.5-Sonnet) to generate keywords and explanations for each demonstration, which are then manually reviewed and refined (See Appendix \ref{appendix:details}).

\textbf{Configuration.} In our experiments, we treat the LLMs as black-box generators and only need to set a few key interface parameters. We maintain consistent settings across all LLMs, employing greedy decoding for output generation. The maximum output length is uniformly set to 2048 tokens. 
Specifically, the LLMs accessed via APIs do not support Beam Search. Thus, we only implement Beam Search for Llama-3.1-70B-Instruct and Mixtral-8×22B-Instruct-v0.1. Due to resource limitation, we compare SelfEvolve using GPT-3.5-turbo following the original paper~\citep{jiang2023selfevolve} and additionally use two open-sourced LLMs (Llama-3.1 and Mixtral-8x22B).

\section{Experimental Results}

\subsection{Main Results}
\label{sec_human}
Table \ref{tab:main} presents the performance of \SEK and the selected baselines across five representative LLMs on Humaneval(+), MBPP(+) and APPS of different difficulty levels. To be noted, the Default results of Mixtral-8×22B-Instruct-v0.1 and DeepSeekCoder-V2-Instruct on Humaneval(+) and MBPP(+) are from the official leaderboard of the EvalPlus~\citep{liu2024your}. However, as the other three LLMs are not in this leaderboard, we adhere to the EvalPlus framework to obtain their results.

Overall, \SEK substantially improves code generation performance, achieving notable gains across various LLMs and datasets. We observe that \SEK achieves greater performance improvements on HumanEval(+) and APPS than MBPP(+). For instance, on HumanEval, \SEK demonstrates an absolute average performance improvement of 4.4\% over the Default, whereas, it achieves an improvement of 1.8\% on MBPP. This may be because the problems in HumanEval(+) and APPS are more complex than those in MBPP, and simple problems are easy to understand and alleviate the need to extract and explain keywords. As shown in Table~\ref{tab_benchmark}, the average number of tokens per problem is 26.1 for MBPP, while those numbers are 67.7 and more than 257.3 for HumanEval(+) and APPS. These results may indicate that \SEK can better improve LLMs' problem-solving capabilities on relatively complex problems than on simple problems.

\begin{table}[t]
\centering
\resizebox{\linewidth}{!}{
\begin{tabular}{@{}cccccccccc@{}}
\toprule
\textbf{Model} & \textbf{Method} & \textbf{HumanEval} & \textbf{HumanEval+} & \textbf{MBPP} & \textbf{MBPP+} & \begin{tabular}[c]{@{}c@{}}\textbf{APPS}\\ \textbf{Introductory}\end{tabular} & \begin{tabular}[c]{@{}c@{}}\textbf{APPS}\\ \textbf{Interview}\end{tabular} & \begin{tabular}[c]{@{}c@{}}\textbf{APPS}\\ \textbf{Competition}\end{tabular} & \textbf{Average} \\ \midrule
\multirow{5}{*}{Llama-3.1-70B-Instruct} & Default & 78.0 & 73.8 & 87.6 & 70.9 & 50.0 & 15.0 & 5.0 & 54.3 \\
 & Beam Search & 79.3 & 74.4 & 87.8 & 70.9 & 55.0 & 16.1 & 5.0 & 55.5 \\
 & CoT & 79.9 & 74.4 & 87.0 & \textbf{71.7} & 43.3 & 16.6 & 6.7 & 54.2 \\
 & SelfEvolve & 81.7 & 75.6 & 85.4 & 70.4 & 50.0 & 15.5 & \textbf{8.3} & 55.3 \\ \cmidrule(l){2-10} 
 & SEK & \textbf{84.8} & \textbf{79.3} & \textbf{88.4} & 71.2 & \textbf{61.7} & \textbf{20.0} & \textbf{8.3} & \textbf{59.1} \\ \midrule
\multirow{5}{*}{Mixtral-8×22B-Instruct-v0.1} & Default & 76.2 & 72.0 & 73.8 & 64.3 & 28.3 & 7.7 & 1.6 & 46.3 \\
 & Beam Search & 78.7 & 73.2 & \textbf{81.2} & \textbf{70.6} & \textbf{33.3} & 8.8 & \textbf{6.6} & 50.3 \\
 & CoT & 72.0 & 65.9 & 78.0 & 68.0 & 31.6 & 3.8 & 5.0 & 46.3 \\
 & SelfEvolve & 56.7 & 50.0 & 68.5 & 60.1 & \textbf{33.3} & 7.2 & 5.0 & 40.1 \\ \cmidrule(l){2-10} 
 & SEK & \textbf{81.1} & \textbf{75.6} & 79.1 & 66.9 & \textbf{33.3} & \textbf{10.0} & \textbf{6.6} & \textbf{50.4} \\ \midrule
\multirow{4}{*}{\begin{tabular}[c]{@{}c@{}}GPT-3.5-turbo\\ (API)\end{tabular}} & Default & 72.6 & 67.7 & \textbf{84.1} & 71.2 & 46.6 & 18.3 & 0.0 & 51.5 \\
 & CoT & 58.5 & 54.9 & \textbf{84.1} & 68.8 & 41.6 & 17.2 & 1.6 & 46.7 \\
 & SelfEvolve & 73.2 & 67.7 & 82.3 & 66.7 & 45.0 & 19.4 & 1.6 & 51.8 \\ \cmidrule(l){2-10} 
 & SEK & \textbf{75.6} & \textbf{69.5} & \textbf{84.1} & \textbf{72.5} & \textbf{53.3} & \textbf{20.6} & \textbf{5.0} & \textbf{54.4} \\ \midrule
\multirow{3}{*}{\begin{tabular}[c]{@{}c@{}}GPT-4o-mini\\ (API)\end{tabular}} & Default & \textbf{87.8} & 84.1 & 85.7 & 72.8 & 53.3 & 31.6 & 11.6 & 61.0 \\
 & CoT & 87.2 & 84.1 & \textbf{88.1} & 73.3 & 50.0 & 33.8 & 11.6 & 61.2 \\ \cmidrule(l){2-10} 
 & SEK & 87.2 & \textbf{84.1} & 87.8 & \textbf{74.1} & \textbf{58.3} & \textbf{35.0} & \textbf{13.3} & \textbf{62.8} \\ \midrule
\multirow{3}{*}{\begin{tabular}[c]{@{}c@{}}DeepSeekCoder-V2-Instruct\\ (API)\end{tabular}} & Default & 85.4 & 82.3 & 89.4 & 75.1 & 70.0 & 36.1 & 10.0 & 64.0 \\
 & CoT & 88.4 & 82.3 & \textbf{90.5} & 75.4 & 60.0 & 40.5 & 10.0 & 63.9 \\ \cmidrule(l){2-10} 
 & SEK & \textbf{93.3} & \textbf{85.4} & 90.2 & \textbf{76.2} & \textbf{75.0} & \textbf{41.1} & \textbf{13.3} & \textbf{67.8} \\ \bottomrule
\end{tabular}
}
\caption{Pass@1 (\%) results of \SEK and baseline methods on HumanEval(+), MBPP(+) and APPS of different difficulty levels. \textbf{Bold} numbers indicate the best-performing baseline for each model.}
\label{tab:main}
\end{table}

We first discuss the performance on HumanEval(+) and APPS.
These benchmarks are relatively complex compared to MBPP, and better demonstrate the effectiveness of \SEK. 
\SEK consistently outperforms Default across most LLMs. For instance, \SEK
achieves average absolute improvements of 6.7\%, 3.6\%, and 3.7\% on APPS-Introductory, APPS-Interview, and APPS-Competition, respectively. 
However, GPT-4o-mini is an exception, which experiences a slight performance decline on Humaneval(+). This may be because the built-in prudence of GPT-4o-mini~\citep{huang2024reasoning} makes it tend to select more generic keywords, and such generic keywords fail to help LLMs understand low-frequency terms in the problem description. This conjecture is further underpinned by an observation that CoT similarly fails to enhance GPT-4o-mini's performance. The consistent improvements of \SEK across most LLMs highlight its effectiveness in enhancing the problem-solving capabilities of LLMs.

Compared to Beam Search, which also invokes LLMs twice, \SEK shows notable performance improvements. For instance, on Humaneval and Humaneval+, \SEK achieves absolute average improvements of 4.0\% and 3.7\%, respectively, over Beam Search. These can be attributed to \SEK's unique technique: appending the problem's critical parts to the end, enabling LLMs to focus on and comprehend these key concepts. In contrast, Beam Search merely expands the search space without gaining a deep understanding of the problem, leading to lower diversity in outputs~\citep{li2016mutual}. Consequently, it cannot enhance problem-solving capabilities in a targeted manner like \SEK (See Appendix~\ref{appendix:beam} for different cases).

Compared to CoT and SelfEvolve, \SEK demonstrates a notable and consistent performance advantage. For instance, on Humaneval and Humaneval+, \SEK achieves absolute average performance improvements of 7.2\% and 6.5\% over CoT. In contrast, the performance of CoT and SelfEvolve are inconsistent, sometimes even lower than Default. For instance, with Mixtral-8×22B-Instruct-v0.1, SelfEvolve's performance on APPS-Interview is 0.5\% lower than Default. 
The unstable performance of CoT can be attributed to its inherent unsuitability for generation tasks~\citep{sprague2024cot}. Similar phenomena have been observed in prior work~\citep{wang2024rat,zhang2024ratt,li2023structured,jiang2023self}. While both SelfEvolve and \SEK utilize LLMs to extract relevant knowledge from problem descriptions, they differ in the types of extracted knowledge. SEK focuses on low-frequency keywords, which are more difficult to be mapped to code implementation. This enables SEK to effectively fill the knowledge gaps during code generation. In contrast, SelfEvolve tends to merely restate the complete problem description for problems in code generation benchmarks. As a result, it is less effective in addressing the specific knowledge gaps that hinder LLM performance in code generation.

We then discuss the performance on MBPP(+), a relatively simple benchmark. \SEK surpasses SelfEvolve and Default across most LLMs, further demonstrating \SEK's effectiveness. For instance, when applied to Llama-3.1-70B-Instruct, \SEK achieves performance improvements of 3.0\% and 0.8\% over SelfEvolve on MBPP and MBPP+, respectively.

\subsection{Discussion}\label{sec:discussion}
We conduct additional experiments to comprehensively evaluate \SEK's performance and robustness.

\begin{figure*}[t]
\centering
\subfigure[HumanEval]{
    \label{fig:ablation1}
    \includegraphics[width=0.31\textwidth]{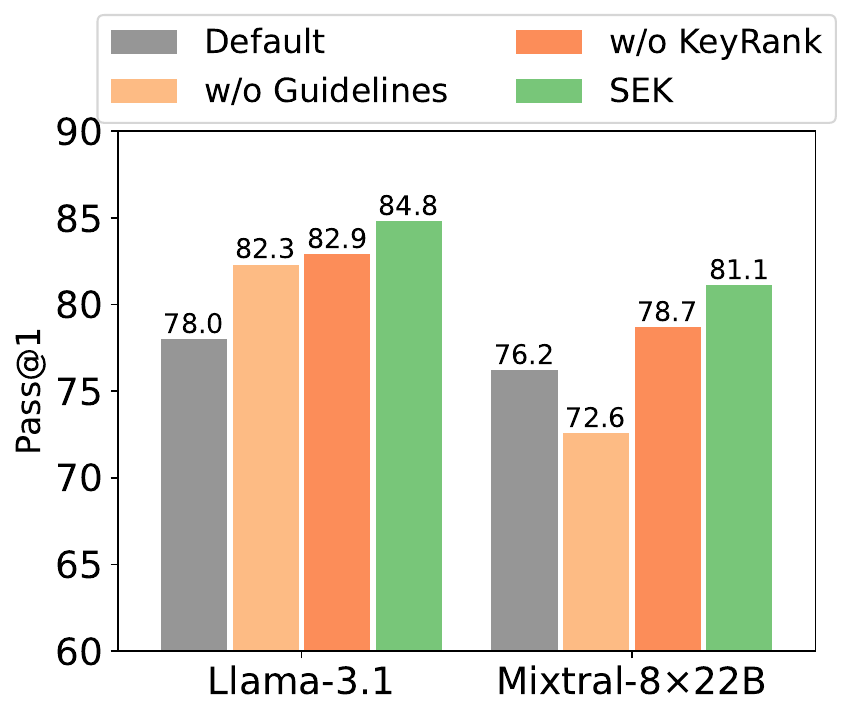}
}%
\subfigure[HumanEval+]{
    \label{fig:ablation2}
    \includegraphics[width=0.31\textwidth]{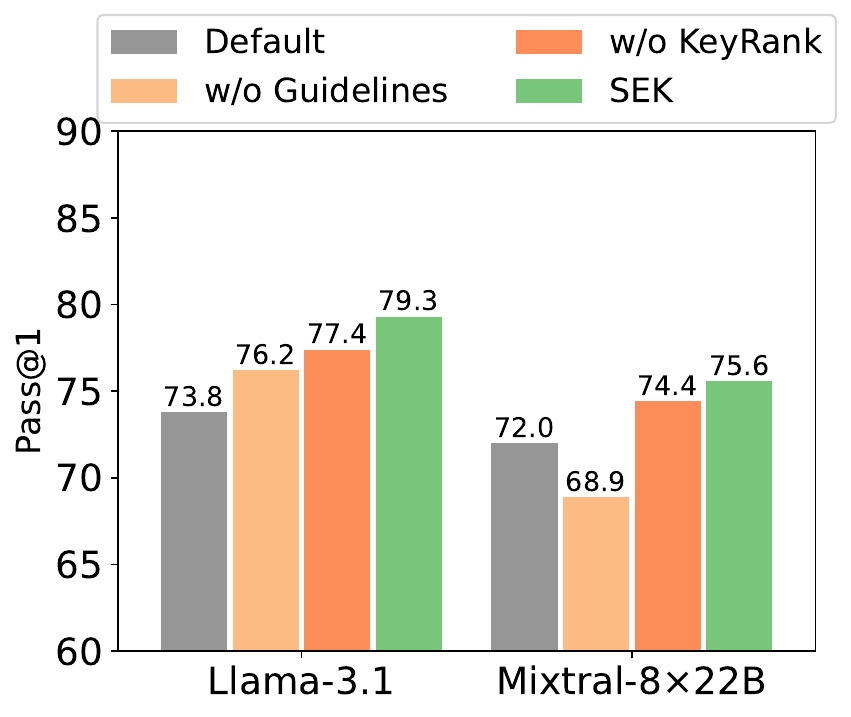}
}%
\subfigure[HumanEval+]{
    \label{fig:robustness}
    \includegraphics[width=0.325\textwidth]{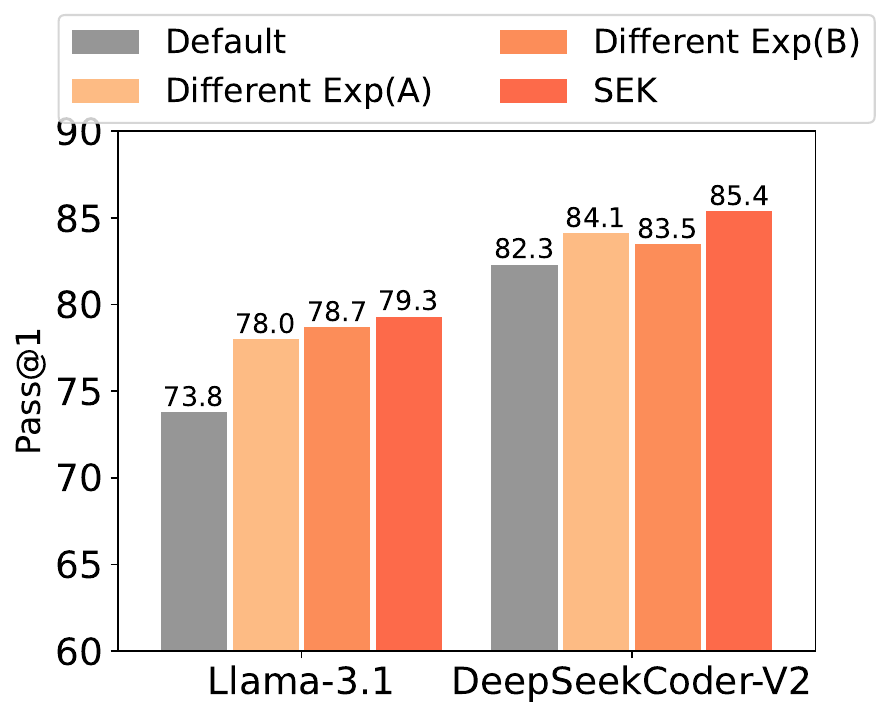}
}
    \caption{(a-b) Ablation experiments on the Humaneval(+) benchmarks with two LLMs. (c) Different explanations of Demonstrations on Humaneval+ with two LLMs.}
\end{figure*}

\begin{table}[t]
\centering
\resizebox{0.8\linewidth}{!}{
\begin{tabular}{@{}clcc@{}}
\toprule
\textbf{Model} & \textbf{Method} & \multicolumn{1}{l}{\textbf{Humaneval}} & \multicolumn{1}{l}{\textbf{Humaneval+}} \\ \midrule
\multirow{3}{*}{Llama-3.1-70B-Instruct} & Default & 78.0 & 73.8 \\
 & \SEK (corpus = APPS training set) & 84.1 & 78.7 \\
 & \SEK (corpus = Python subset of eval-codealpaca-v1) & \textbf{84.8} & \textbf{79.3} \\ \midrule
\multirow{3}{*}{DeepSeekCoder-V2-Instruct} & Default & 85.4 & 82.3 \\
 & \SEK (corpus = APPS training set) & 90.9 & \textbf{85.4} \\
 & \SEK (corpus = Python subset of eval-codealpaca-v1) & \textbf{93.3} & \textbf{85.4} \\ \bottomrule
\end{tabular}
}
\caption{\SEK works under different corpus for Humaneval(+).}
\label{tab:robust_corpus}
\end{table}

\textbf{Guidelines in the prompt for KeyExtract \& Explain provide essential guidance for LLMs, and KeyRank effectively prioritizes keywords}. Our ablation studies confirm that both guidelines and KeyRank play crucial roles in enhancing performance. As shown in Figure~\ref{fig:ablation1}-\ref{fig:ablation2}, We evaluate Llama-3.1 and Mixtral-8×22B on Humaneval (+).  Removing either the guidelines or the KeyRank module results in performance degradation. For instance, removing the KeyRank module results in performance decreases of 2.4\% and 1.2\% on HumanEval and HumanEval+, respectively, for Mixtral-8×22B-Instruct-v0.1. Moreover, removing each guideline from the prompt individually also results in performance degradation in most cases (See Appendix~\ref{append:guide}). It is worth mentioning that even without KeyRank, \SEK remains superior to the Default baseline. For instance, without KeyRank module, Mixtral-8×22B-Instruct-v0.1 shows a 2.5\% improvement on HumanEval compared to the Default, underscoring the strength of \SEK's core mechanisms.

\textbf{\SEK demonstrates robustness to variations in demonstrations, and the corpus used in KeyRank}. To show its performance is not tied to a fixed set of keyword explanations within the demonstrations used in KeyExtract \& Explain, We conduct experiments using two additional sets of keyword explanations randomly generated from the same LLM (i.e., Claude-3.5-Sonnet). As shown in Figure~\ref{fig:robustness}, although there is performance variance among different keyword explanations, as would be expected when using exemplar-based prompting ~\citep{gao2021making,min2022rethinking,reynolds2021prompt}, the three sets of keyword explanations consistently outperform the Default. Additionally, to evaluate the robustness to the corpus used in KeyRank, we employ select different corpus, as shown in Table~\ref{tab:robust_corpus}. We observe that using  \SEK with Llama-3.1-70B-Instruct still shows a 6.1\% absolute improvement on Humaneval compared to Default. These results demonstrate the robustness of \SEK. 

\textbf{\SEK enhances the model’s focus on core keywords in the problem description} (See Appendix~\ref{appendix:attention}). Using a visualization tool, we analyze \SEK's behaviors from the perspective of attention distribution. We select a simple problem, i.e., ``Write a function to find the nth nonagonal number'', choosing the keyword ``nonagonal'' with its explanation for detailed analysis. By comparing the attention distribution in the Default and \SEK settings, we observe that \SEK help the LLM allocate more attention to the keyword and its explanation. This indicates the way \SEK uses to enrich the prompt can help LLMs better focus on the key concepts in the problem description, leading to improved code generation.

\subsection{Case Study}
\begin{figure}[t]
\centering
\includegraphics[width=\textwidth]{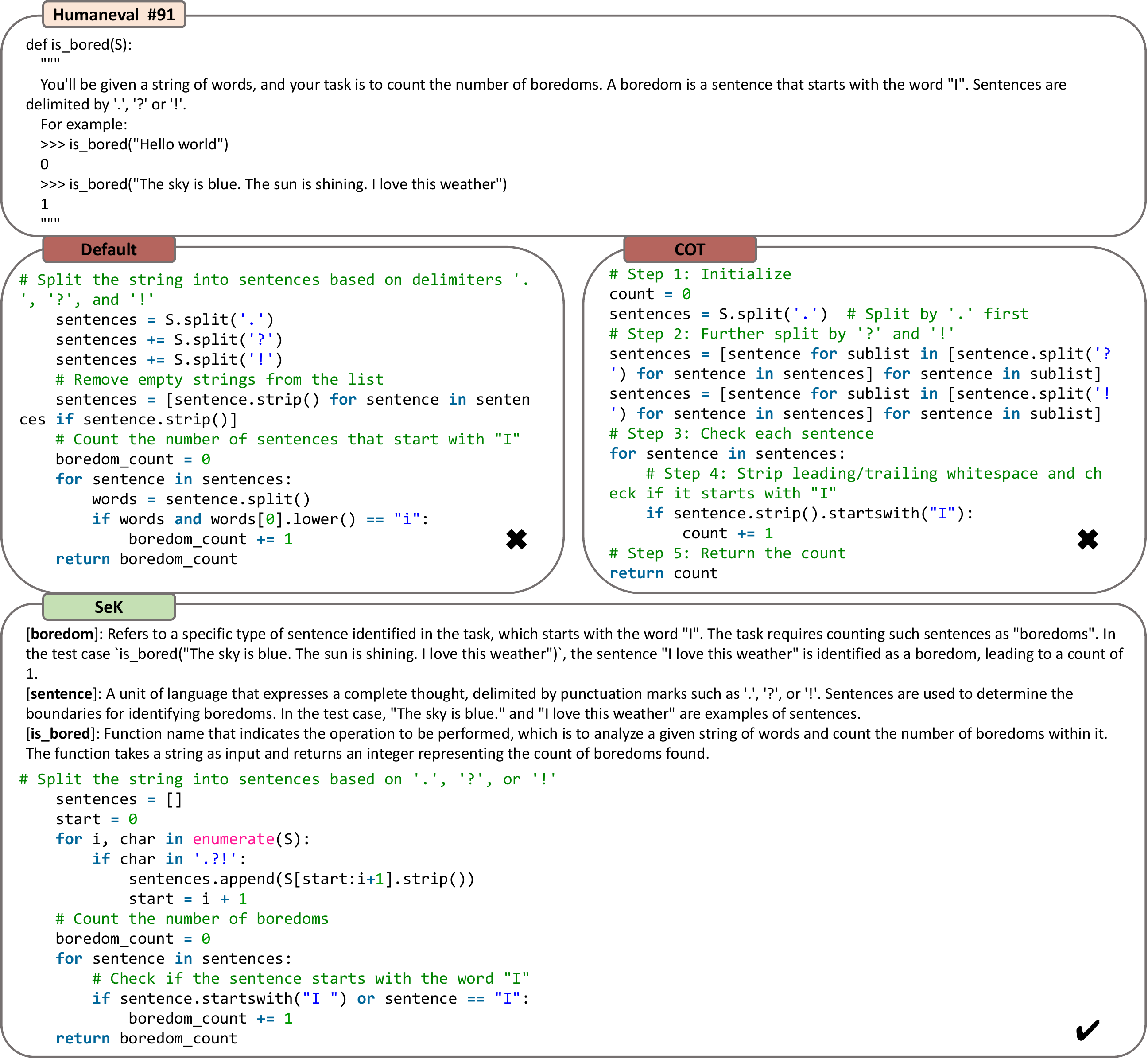}
\caption{A real case from HumanEval generated by two baselines and \SEK.}
\label{fig_human}
\end{figure}

To further evaluate the effectiveness of \SEK, we conduct a qualitative analysis. As shown in Figures \ref{fig_human}, we select one representative sample from HumanEval, use DeepSeek-Coder-V2-Instruct as the base model, and compare the outputs of \SEK with Default and CoT. See Appendix \ref{append:case} for more details.

The problem requires finding \textit{boredom} from a string of words. A \textit{boredom} refers to a sentence that begins with words starting with ``I". Default fails at the very beginning by incorrectly segmenting \textit{sentences}, possibly due to an inadequate understanding of the concept \textit{sentence}. Although CoT correctly segments \textit{sentences} through chain-of-thought reasoning, it errs when searching for \textit{boredom}, mistakenly seeking \textit{sentences} starting with ``I" instead of \textit{boredom}, indicating an incomplete understanding of the \textit{boredom} concept. In contrast, \SEK correctly identifies key concepts in the problem: \textit{sentence}, \textit{boredom}, and the method name \textit{is\_bored}. In their explanations, \SEK precisely explains these key concepts by incorporating the provided test cases.

Through reranking, \SEK helps the LLM understand the problem according to the importance of concepts: The LLM first comprehends the concept of \textit{boredom}, then the \textit{sentence}, and finally the \textit{is\_bored}. This precise comprehension and organization of key concepts enable \SEK to grasp the core elements of the problem, thereby generating correct code solutions.

\section{Related Work}
\textbf{LLM-based code generation}: Recent advancements in LLMs have significantly improved code generation capabilities. Models like CodeGen~\citep{nijkamp2022codegen}, StarCoder~\citep{li2023starcoder}, and GPT series~\citep{black2022gpt,chen2021evaluating} have demonstrated remarkable performance in translating natural language descriptions into code snippets. These models primarily use decoder-only architectures and next-token prediction for pre-training. A subset, including CodeT5 ~\citep{wang2021codet5} and PLBART~\citep{ahmad2021unified}, employs encoder-decoder architectures. Our work builds upon these foundations, focusing on enhancing LLMs' problem-solving capabilities without additional training.

\textbf{Prompting techniques for code generation}: Prompting techniques for code generation can be broadly categorized into three types: The first type utilizes external knowledge to enhance LLMs' understanding of coding problems or intermediate outputs~\citep{mu2023clarifygpt,nashid2023retrieval,zhong2024debug}. For example, 
CEDAR~\citep{nashid2023retrieval} retrieves relevant code examples from an external knowledge base to help LLMs understand task requirements. The second type relies solely on LLMs' inherent capabilities, using prompt design to guide LLMs in generating code snippets that meet specific requirements~\citep{wei2022chain,wang2022self,yao2024tree}. For instance, Chain of Thought~\citep{wei2022chain} employs a step-by-step, chain-of-thought style prompt to guide LLMs in producing correct results. 
The third type integrates the previous two types, leveraging both external knowledge and the LLM's inherent knowledge to solve coding problems~\citep{chen2023teaching,jiang2023selfevolve,tian2023test,chen-etal-2024-jumpcoder,chen2024b4}. For example, Self-Debug~\citep{chen2023teaching} uses the code execution results or the code explanations generated by the LLM itself to debug the incorrect code multiple times. 
\SEK belongs to the second category.
Different from other methods, it focuses on improving LLMs' comprehension of the problem by identifying and explaining the key concepts in the problem description with LLMs themselves.

\textbf{Keyword extraction}: Keyword extraction methods have evolved from traditional statistical~\citep{sparck1972statistical,el2009kp,florescu2017new,rose2010automatic} and graph-based approaches~\citep{mihalcea2004textrank,wan2008single,gollapalli2014extracting,grineva2009extracting} to more advanced techniques leveraging language models~\citep{mahata2018key2vec,bennani2018simple,sun2020sifrank,arora2017simple}. Recent works like AttentionRank~\citep{ding2021attentionrank} and LLM-TAKE~\citep{maragheh2023llm} use self-attention mechanisms and language models to identify significant keywords. Our work extends this concept to the domain of code generation, using LLMs to extract and explain problem-specific keywords to enhance code solution generation.

\section{Conclusion and Limitations}

In this work, we propose \SEK, a simple yet effective method to enhance the code generation capabilities of LLMs. \SEK leverages the LLM to extract and explain keywords from the problem description, followed by ranking them based on their frequency. Through extensive experiments, we demonstrate that \SEK facilitates LLMs in capturing and clarifying key concepts within problems, thereby generating more accurate code solutions.

One limitation of \SEK is that the two-stage invocation process of \SEK incurs additional computational overhead. Future work could explore compressing the process into one invocation. In addition, keywords are extracted and explained by LLMs, of which the quality cannot be guaranteed due to the hallucinations of LLMs~\citep{ji2023survey}. Mitigating this requires enhancing the factual accuracy of LLMs~\citep{mitchell2022memory,tang2023towards} and proposing effective approaches for detecting factual errors~\citep{chen2024complex,min2023factscore}.

\bibliography{iclr2024_conference}

\begin{thebibliography}{80}
\providecommand{\natexlab}[1]{#1}
\providecommand{\url}[1]{\texttt{#1}}
\expandafter\ifx\csname urlstyle\endcsname\relax
  \providecommand{\doi}[1]{doi: #1}\else
  \providecommand{\doi}{doi: \begingroup \urlstyle{rm}\Url}\fi

\bibitem[Ahmad et~al.(2021)Ahmad, Chakraborty, Ray, and Chang]{ahmad2021unified}
WU~Ahmad, S~Chakraborty, B~Ray, and KW~Chang.
\newblock Unified pre-training for program understanding and generation.
\newblock In \emph{Proceedings of the 2021 Conference of the North American Chapter of the Association for Computational Linguistics: Human Language Technologies}, 2021.

\bibitem[AI@Meta(2024)]{llama3modelcard}
AI@Meta.
\newblock Llama 3 model card.
\newblock 2024.
\newblock URL \url{https://github.com/meta-llama/llama3/blob/main/MODEL_CARD.md}.

\bibitem[Arora et~al.(2017)Arora, Liang, and Ma]{arora2017simple}
Sanjeev Arora, Yingyu Liang, and Tengyu Ma.
\newblock A simple but tough-to-beat baseline for sentence embeddings.
\newblock In \emph{International conference on learning representations}, 2017.

\bibitem[Austin et~al.(2021)Austin, Odena, Nye, Bosma, Michalewski, Dohan, Jiang, Cai, Terry, Le, et~al.]{austin2021program}
Jacob Austin, Augustus Odena, Maxwell Nye, Maarten Bosma, Henryk Michalewski, David Dohan, Ellen Jiang, Carrie Cai, Michael Terry, Quoc Le, et~al.
\newblock Program synthesis with large language models.
\newblock \emph{arXiv preprint arXiv:2108.07732}, 2021.

\bibitem[Bennani-Smires et~al.(2018)Bennani-Smires, Musat, Hossmann, Baeriswyl, and Jaggi]{bennani2018simple}
Kamil Bennani-Smires, Claudiu-Cristian Musat, Andreea Hossmann, Michael Baeriswyl, and Martin Jaggi.
\newblock Simple unsupervised keyphrase extraction using sentence embeddings.
\newblock In \emph{Proceedings of the 22nd Conference on Computational Natural Language Learning}, pp.\  221--229, 2018.

\bibitem[Black et~al.(2022)Black, Biderman, Hallahan, Anthony, Gao, Golding, He, Leahy, McDonell, Phang, Pieler, Prashanth, Purohit, Reynolds, Tow, Wang, and Weinbach]{black2022gpt}
Sidney Black, Stella Biderman, Eric Hallahan, Quentin Anthony, Leo Gao, Laurence Golding, Horace He, Connor Leahy, Kyle McDonell, Jason Phang, Michael Pieler, Usvsn~Sai Prashanth, Shivanshu Purohit, Laria Reynolds, Jonathan Tow, Ben Wang, and Samuel Weinbach.
\newblock Gpt-neox-20b: An open-source autoregressive language model.
\newblock pp.\  95--136, May 2022.
\newblock \doi{10.18653/v1/2022.bigscience-1.9}.
\newblock URL \url{https://aclanthology.org/2022.bigscience-1.9}.

\bibitem[Chen et~al.(2023{\natexlab{a}})Chen, Scheurer, Korbak, Campos, Chan, Bowman, Cho, and Perez]{chen2023improving}
Angelica Chen, J{\'e}r{\'e}my Scheurer, Tomasz Korbak, Jon~Ander Campos, Jun~Shern Chan, Samuel~R Bowman, Kyunghyun Cho, and Ethan Perez.
\newblock Improving code generation by training with natural language feedback.
\newblock \emph{arXiv preprint arXiv:2303.16749}, 2023{\natexlab{a}}.

\bibitem[Chen et~al.(2023{\natexlab{b}})Chen, Zhang, Nguyen, Zan, Lin, Lou, and Chen]{chencodet}
Bei Chen, Fengji Zhang, Anh Nguyen, Daoguang Zan, Zeqi Lin, Jian{-}Guang Lou, and Weizhu Chen.
\newblock Codet: Code generation with generated tests.
\newblock In \emph{The Eleventh International Conference on Learning Representations, {ICLR} 2023, Kigali, Rwanda, May 1-5, 2023}. OpenReview.net, 2023{\natexlab{b}}.
\newblock URL \url{https://openreview.net/forum?id=ktrw68Cmu9c}.

\bibitem[Chen et~al.(2024{\natexlab{a}})Chen, Kim, Sriram, Durrett, and Choi]{chen2024complex}
Jifan Chen, Grace Kim, Aniruddh Sriram, Greg Durrett, and Eunsol Choi.
\newblock Complex claim verification with evidence retrieved in the wild.
\newblock In \emph{Proceedings of the 2024 Conference of the North American Chapter of the Association for Computational Linguistics: Human Language Technologies (Volume 1: Long Papers)}, pp.\  3569--3587, 2024{\natexlab{a}}.

\bibitem[Chen et~al.(2021)Chen, Tworek, Jun, Yuan, Pinto, Kaplan, Edwards, Burda, Joseph, Brockman, et~al.]{chen2021evaluating}
Mark Chen, Jerry Tworek, Heewoo Jun, Qiming Yuan, Henrique Ponde De~Oliveira Pinto, Jared Kaplan, Harri Edwards, Yuri Burda, Nicholas Joseph, Greg Brockman, et~al.
\newblock Evaluating large language models trained on code.
\newblock \emph{arXiv preprint arXiv:2107.03374}, 2021.

\bibitem[Chen et~al.(2024{\natexlab{b}})Chen, Liu, Tao, Hong, Lo, Xia, and Sun]{chen2024b4}
Mouxiang Chen, Zhongxin Liu, He~Tao, Yusu Hong, David Lo, Xin Xia, and Jianling Sun.
\newblock B4: Towards optimal assessment of plausible code solutions with plausible tests.
\newblock \emph{arXiv preprint arXiv:2409.08692}, 2024{\natexlab{b}}.

\bibitem[Chen et~al.(2024{\natexlab{c}})Chen, Tian, Liu, Ren, and Sun]{chen-etal-2024-jumpcoder}
Mouxiang Chen, Hao Tian, Zhongxin Liu, Xiaoxue Ren, and Jianling Sun.
\newblock {J}ump{C}oder: Go beyond autoregressive coder via online modification.
\newblock In Lun-Wei Ku, Andre Martins, and Vivek Srikumar (eds.), \emph{Proceedings of the 62nd Annual Meeting of the Association for Computational Linguistics (Volume 1: Long Papers)}, pp.\  11500--11520, Bangkok, Thailand, August 2024{\natexlab{c}}. Association for Computational Linguistics.
\newblock \doi{10.18653/v1/2024.acl-long.619}.
\newblock URL \url{https://aclanthology.org/2024.acl-long.619}.

\bibitem[Chen et~al.(2024{\natexlab{d}})Chen, Qin, Jiang, and Choi]{chen2024large}
Ruirui Chen, Chengwei Qin, Weifeng Jiang, and Dongkyu Choi.
\newblock Is a large language model a good annotator for event extraction?
\newblock In \emph{Proceedings of the AAAI Conference on Artificial Intelligence}, volume~38, pp.\  17772--17780, 2024{\natexlab{d}}.

\bibitem[Chen et~al.(2023{\natexlab{c}})Chen, Lin, Schaerli, and Zhou]{chen2023teaching}
Xinyun Chen, Maxwell Lin, Nathanael Schaerli, and Denny Zhou.
\newblock Teaching large language models to self-debug.
\newblock In \emph{The 61st Annual Meeting Of The Association For Computational Linguistics}, 2023{\natexlab{c}}.

\bibitem[Ding \& Luo(2021)Ding and Luo]{ding2021attentionrank}
Haoran Ding and Xiao Luo.
\newblock Attentionrank: Unsupervised keyphrase extraction using self and cross attentions.
\newblock In \emph{Proceedings of the 2021 Conference on Empirical Methods in Natural Language Processing}, pp.\  1919--1928, 2021.

\bibitem[Dong et~al.(2023)Dong, Jiang, Jin, and Li]{dongself}
Yihong Dong, Xue Jiang, Zhi Jin, and Ge~Li.
\newblock Self-collaboration code generation via chatgpt.
\newblock \emph{ACM Transactions on Software Engineering and Methodology}, 2023.

\bibitem[Dubey \& Abhinav~Jauhri(2024)Dubey and Abhinav~Jauhri]{dubey2024llama3herdmodels}
Abhimanyu Dubey and et~al. Abhinav~Jauhri.
\newblock The llama 3 herd of models, 2024.
\newblock URL \url{https://arxiv.org/abs/2407.21783}.

\bibitem[Dubois et~al.(2024)Dubois, Li, Taori, Zhang, Gulrajani, Ba, Guestrin, Liang, and Hashimoto]{dubois2024alpacafarm}
Yann Dubois, Chen~Xuechen Li, Rohan Taori, Tianyi Zhang, Ishaan Gulrajani, Jimmy Ba, Carlos Guestrin, Percy~S Liang, and Tatsunori~B Hashimoto.
\newblock Alpacafarm: A simulation framework for methods that learn from human feedback.
\newblock \emph{Advances in Neural Information Processing Systems}, 36, 2024.

\bibitem[El-Beltagy \& Rafea(2009)El-Beltagy and Rafea]{el2009kp}
Samhaa~R El-Beltagy and Ahmed Rafea.
\newblock Kp-miner: A keyphrase extraction system for english and arabic documents.
\newblock \emph{Information systems}, 34\penalty0 (1):\penalty0 132--144, 2009.

\bibitem[Fan et~al.(2024)Fan, Li, Nag, Fang, Biswas, Xu, and Achan]{fan2024prompt}
Zezhong Fan, Xiaohan Li, Kaushiki Nag, Chenhao Fang, Topojoy Biswas, Jianpeng Xu, and Kannan Achan.
\newblock Prompt optimizer of text-to-image diffusion models for abstract concept understanding.
\newblock In \emph{Companion Proceedings of the ACM on Web Conference 2024}, pp.\  1530--1537, 2024.

\bibitem[Fang et~al.(2024)Fang, Li, Fan, Xu, Nag, Korpeoglu, Kumar, and Achan]{fang2024llm}
Chenhao Fang, Xiaohan Li, Zezhong Fan, Jianpeng Xu, Kaushiki Nag, Evren Korpeoglu, Sushant Kumar, and Kannan Achan.
\newblock Llm-ensemble: Optimal large language model ensemble method for e-commerce product attribute value extraction.
\newblock In \emph{Proceedings of the 47th International ACM SIGIR Conference on Research and Development in Information Retrieval}, pp.\  2910--2914, 2024.

\bibitem[Florescu \& Caragea(2017)Florescu and Caragea]{florescu2017new}
Corina Florescu and Cornelia Caragea.
\newblock A new scheme for scoring phrases in unsupervised keyphrase extraction.
\newblock In \emph{Advances in Information Retrieval: 39th European Conference on IR Research, ECIR 2017, Aberdeen, UK, April 8-13, 2017, Proceedings 39}, pp.\  477--483. Springer, 2017.

\bibitem[Gao et~al.(2021)Gao, Fisch, and Chen]{gao2021making}
Tianyu Gao, Adam Fisch, and Danqi Chen.
\newblock Making pre-trained language models better few-shot learners.
\newblock In \emph{Proceedings of the 59th Annual Meeting of the Association for Computational Linguistics and the 11th International Joint Conference on Natural Language Processing (Volume 1: Long Papers)}. Association for Computational Linguistics, 2021.

\bibitem[Gollapalli \& Caragea(2014)Gollapalli and Caragea]{gollapalli2014extracting}
Sujatha~Das Gollapalli and Cornelia Caragea.
\newblock Extracting keyphrases from research papers using citation networks.
\newblock In \emph{Proceedings of the AAAI conference on artificial intelligence}, volume~28, 2014.

\bibitem[Grineva et~al.(2009)Grineva, Grinev, and Lizorkin]{grineva2009extracting}
Maria Grineva, Maxim Grinev, and Dmitry Lizorkin.
\newblock Extracting key terms from noisy and multitheme documents.
\newblock In \emph{Proceedings of the 18th international conference on World wide web}, pp.\  661--670, 2009.

\bibitem[Hendrycks et~al.(2021)Hendrycks, Basart, Kadavath, Mazeika, Arora, Guo, Burns, Puranik, He, Song, et~al.]{hendrycks2measuring}
Dan Hendrycks, Steven Basart, Saurav Kadavath, Mantas Mazeika, Akul Arora, Ethan Guo, Collin Burns, Samir Puranik, Horace He, Dawn Song, et~al.
\newblock Measuring coding challenge competence with apps.
\newblock In \emph{Thirty-fifth Conference on Neural Information Processing Systems Datasets and Benchmarks Track (Round 2)}, 2021.

\bibitem[Huang et~al.(2024{\natexlab{a}})Huang, Gu, Hu, Li, Li, and Xu]{huang2024reasoning}
Sirui Huang, Yanggan Gu, Xuming Hu, Zhonghao Li, Qing Li, and Guandong Xu.
\newblock Reasoning factual knowledge in structured data with large language models.
\newblock \emph{arXiv preprint arXiv:2408.12188}, 2024{\natexlab{a}}.

\bibitem[Huang et~al.(2024{\natexlab{b}})Huang, Sun, Jin, Li, and Lyu]{huang2024knowledge}
Tao Huang, Zhihong Sun, Zhi Jin, Ge~Li, and Chen Lyu.
\newblock Knowledge-aware code generation with large language models.
\newblock In \emph{Proceedings of the 32nd IEEE/ACM International Conference on Program Comprehension}, pp.\  52--63, 2024{\natexlab{b}}.

\bibitem[Hunt \& Thomas.(2000)Hunt and Thomas.]{andrewthe00}
Andrew Hunt and David Thomas.
\newblock \emph{The pragmatic programmer: from journeyman to master}.
\newblock Addison-Wesley Longman Publishing Co., Inc., 2000.

\bibitem[Inala et~al.(2022)Inala, Wang, Yang, Codas, Encarnaci{\'o}n, Lahiri, Musuvathi, and Gao]{inala2022fault}
Jeevana~Priya Inala, Chenglong Wang, Mei Yang, Andres Codas, Mark Encarnaci{\'o}n, Shuvendu Lahiri, Madanlal Musuvathi, and Jianfeng Gao.
\newblock Fault-aware neural code rankers.
\newblock \emph{Advances in Neural Information Processing Systems}, 35:\penalty0 13419--13432, 2022.

\bibitem[Ji et~al.(2023)Ji, Lee, Frieske, Yu, Su, Xu, Ishii, Bang, Madotto, and Fung]{ji2023survey}
Ziwei Ji, Nayeon Lee, Rita Frieske, Tiezheng Yu, Dan Su, Yan Xu, Etsuko Ishii, Ye~Jin Bang, Andrea Madotto, and Pascale Fung.
\newblock Survey of hallucination in natural language generation.
\newblock \emph{ACM Computing Surveys}, 55\penalty0 (12):\penalty0 1--38, 2023.

\bibitem[Jiang et~al.(2024)Jiang, Sablayrolles, Roux, Mensch, Savary, Bamford, Chaplot, de~las Casas, Hanna, Bressand, Lengyel, Bour, Lample, Lavaud, Saulnier, Lachaux, Stock, Subramanian, Yang, Antoniak, Scao, Gervet, Lavril, Wang, Lacroix, and Sayed]{jiang2024mixtralexperts}
Albert~Q. Jiang, Alexandre Sablayrolles, Antoine Roux, Arthur Mensch, Blanche Savary, Chris Bamford, Devendra~Singh Chaplot, Diego de~las Casas, Emma~Bou Hanna, Florian Bressand, Gianna Lengyel, Guillaume Bour, Guillaume Lample, Lélio~Renard Lavaud, Lucile Saulnier, Marie-Anne Lachaux, Pierre Stock, Sandeep Subramanian, Sophia Yang, Szymon Antoniak, Teven~Le Scao, Théophile Gervet, Thibaut Lavril, Thomas Wang, Timothée Lacroix, and William~El Sayed.
\newblock Mixtral of experts, 2024.
\newblock URL \url{https://arxiv.org/abs/2401.04088}.

\bibitem[Jiang et~al.(2023{\natexlab{a}})Jiang, Wang, and Wang]{jiang2023selfevolve}
Shuyang Jiang, Yuhao Wang, and Yu~Wang.
\newblock Selfevolve: A code evolution framework via large language models.
\newblock \emph{arXiv preprint arXiv:2306.02907}, 2023{\natexlab{a}}.

\bibitem[Jiang et~al.(2023{\natexlab{b}})Jiang, Dong, Wang, Zheng, Shang, Li, Jin, and Jiao]{jiang2023self}
Xue Jiang, Yihong Dong, Lecheng Wang, Fang Zheng, Qiwei Shang, Ge~Li, Zhi Jin, and Wenpin Jiao.
\newblock Self-planning code generation with large language models.
\newblock \emph{ACM Transactions on Software Engineering and Methodology}, 2023{\natexlab{b}}.

\bibitem[Le et~al.(2024)Le, Chen, Saha, Gokul, Sahoo, and Joty]{lecodechain}
Hung Le, Hailin Chen, Amrita Saha, Akash Gokul, Doyen Sahoo, and Shafiq Joty.
\newblock Codechain: Towards modular code generation through chain of self-revisions with representative sub-modules.
\newblock In \emph{The Twelfth International Conference on Learning Representations}, 2024.

\bibitem[Lee et~al.(2023)Lee, Chun, Jeong, and Jung]{lee2023toward}
Wanhae Lee, Minki Chun, Hyeonhak Jeong, and Hyunggu Jung.
\newblock Toward keyword generation through large language models.
\newblock In \emph{Companion Proceedings of the 28th International Conference on Intelligent User Interfaces}, pp.\  37--40, 2023.

\bibitem[Li \& Jurafsky(2016)Li and Jurafsky]{li2016mutual}
Jiwei Li and Dan Jurafsky.
\newblock Mutual information and diverse decoding improve neural machine translation.
\newblock \emph{arXiv preprint arXiv:1601.00372}, 2016.

\bibitem[Li et~al.(2024{\natexlab{a}})Li, Chen, Ren, Cheng, Zhao, Nie, and Wen]{li2024dawn}
Junyi Li, Jie Chen, Ruiyang Ren, Xiaoxue Cheng, Wayne~Xin Zhao, Jian-Yun Nie, and Ji-Rong Wen.
\newblock The dawn after the dark: An empirical study on factuality hallucination in large language models.
\newblock \emph{arXiv preprint arXiv:2401.03205}, 2024{\natexlab{a}}.

\bibitem[Li et~al.(2023)Li, Allal, Zi, Muennighoff, Kocetkov, Mou, Marone, Akiki, Li, Chim, et~al.]{li2023starcoder}
R~Li, LB~Allal, Y~Zi, N~Muennighoff, D~Kocetkov, C~Mou, M~Marone, C~Akiki, J~Li, J~Chim, et~al.
\newblock Starcoder: May the source be with you!
\newblock \emph{Transactions on machine learning research}, 2023.

\bibitem[Li et~al.(2024{\natexlab{b}})Li, Zhang, Do, Yue, and Chen]{li2024long}
Tianle Li, Ge~Zhang, Quy~Duc Do, Xiang Yue, and Wenhu Chen.
\newblock Long-context llms struggle with long in-context learning.
\newblock \emph{arXiv preprint arXiv:2404.02060}, 2024{\natexlab{b}}.

\bibitem[Liu et~al.(2024)Liu, Xia, Wang, and Zhang]{liu2024your}
Jiawei Liu, Chunqiu~Steven Xia, Yuyao Wang, and Lingming Zhang.
\newblock Is your code generated by chatgpt really correct? rigorous evaluation of large language models for code generation.
\newblock \emph{Advances in Neural Information Processing Systems}, 36, 2024.

\bibitem[Luo et~al.(2023)Luo, Xu, Zhao, Sun, Geng, Hu, Tao, Ma, Lin, and Jiang]{luo2023wizardcoder}
Ziyang Luo, Can Xu, Pu~Zhao, Qingfeng Sun, Xiubo Geng, Wenxiang Hu, Chongyang Tao, Jing Ma, Qingwei Lin, and Daxin Jiang.
\newblock Wizardcoder: Empowering code large language models with evol-instruct, 2023.

\bibitem[Luo et~al.(2024)Luo, Xu, Zhao, Sun, Geng, Hu, Tao, Ma, Lin, and Jiang]{li2023structured}
Ziyang Luo, Can Xu, Pu~Zhao, Qingfeng Sun, Xiubo Geng, Wenxiang Hu, Chongyang Tao, Jing Ma, Qingwei Lin, and Daxin Jiang.
\newblock Wizardcoder: Empowering code large language models with evol-instruct.
\newblock \emph{The Twelfth International Conference on Learning Representations, {ICLR} 2024, Vienna, Austria, May 7-11, 2024}, 2024.
\newblock URL \url{https://openreview.net/forum?id=UnUwSIgK5W}.

\bibitem[Mahata et~al.(2018)Mahata, Kuriakose, Shah, and Zimmermann]{mahata2018key2vec}
Debanjan Mahata, John Kuriakose, Rajiv Shah, and Roger Zimmermann.
\newblock Key2vec: Automatic ranked keyphrase extraction from scientific articles using phrase embeddings.
\newblock In \emph{Proceedings of the 2018 Conference of the North American Chapter of the Association for Computational Linguistics: Human Language Technologies, Volume 2 (Short Papers)}, pp.\  634--639, 2018.

\bibitem[Maragheh et~al.(2023)Maragheh, Fang, Irugu, Parikh, Cho, Xu, Sukumar, Patel, Korpeoglu, Kumar, et~al.]{maragheh2023llm}
Reza~Yousefi Maragheh, Chenhao Fang, Charan~Chand Irugu, Parth Parikh, Jason Cho, Jianpeng Xu, Saranyan Sukumar, Malay Patel, Evren Korpeoglu, Sushant Kumar, et~al.
\newblock Llm-take: theme-aware keyword extraction using large language models.
\newblock In \emph{2023 IEEE International Conference on Big Data (BigData)}, pp.\  4318--4324. IEEE, 2023.

\bibitem[Mihalcea \& Tarau(2004)Mihalcea and Tarau]{mihalcea2004textrank}
Rada Mihalcea and Paul Tarau.
\newblock Textrank: Bringing order into text.
\newblock In \emph{Proceedings of the 2004 conference on empirical methods in natural language processing}, pp.\  404--411, 2004.

\bibitem[Min et~al.(2022)Min, Lyu, Holtzman, Artetxe, Lewis, Hajishirzi, and Zettlemoyer]{min2022rethinking}
Sewon Min, Xinxi Lyu, Ari Holtzman, Mikel Artetxe, Mike Lewis, Hannaneh Hajishirzi, and Luke Zettlemoyer.
\newblock Rethinking the role of demonstrations: What makes in-context learning work?
\newblock In \emph{Proceedings of the 2022 Conference on Empirical Methods in Natural Language Processing}, pp.\  11048--11064, 2022.

\bibitem[Min et~al.(2023)Min, Krishna, Lyu, Lewis, Yih, Koh, Iyyer, Zettlemoyer, and Hajishirzi]{min2023factscore}
Sewon Min, Kalpesh Krishna, Xinxi Lyu, Mike Lewis, Wen-tau Yih, Pang~Wei Koh, Mohit Iyyer, Luke Zettlemoyer, and Hannaneh Hajishirzi.
\newblock Factscore: Fine-grained atomic evaluation of factual precision in long form text generation.
\newblock In \emph{The 2023 Conference on Empirical Methods in Natural Language Processing}, 2023.

\bibitem[Mitchell et~al.(2022)Mitchell, Lin, Bosselut, Manning, and Finn]{mitchell2022memory}
Eric Mitchell, Charles Lin, Antoine Bosselut, Christopher~D Manning, and Chelsea Finn.
\newblock Memory-based model editing at scale.
\newblock In \emph{International Conference on Machine Learning}, pp.\  15817--15831. PMLR, 2022.

\bibitem[Mu et~al.(2023)Mu, Shi, Wang, Yu, Zhang, Wang, Liu, and Wang]{mu2023clarifygpt}
Fangwen Mu, Lin Shi, Song Wang, Zhuohao Yu, Binquan Zhang, Chenxue Wang, Shichao Liu, and Qing Wang.
\newblock Clarifygpt: Empowering llm-based code generation with intention clarification.
\newblock \emph{arXiv preprint arXiv:2310.10996}, 2023.

\bibitem[Nashid et~al.(2023)Nashid, Sintaha, and Mesbah]{nashid2023retrieval}
Noor Nashid, Mifta Sintaha, and Ali Mesbah.
\newblock Retrieval-based prompt selection for code-related few-shot learning.
\newblock In \emph{2023 IEEE/ACM 45th International Conference on Software Engineering (ICSE)}, pp.\  2450--2462. IEEE, 2023.

\bibitem[Nijkamp et~al.(2022)Nijkamp, Pang, Hayashi, Tu, Wang, Zhou, Savarese, and Xiong]{nijkamp2022codegen}
Erik Nijkamp, Bo~Pang, Hiroaki Hayashi, Lifu Tu, Huan Wang, Yingbo Zhou, Silvio Savarese, and Caiming Xiong.
\newblock Codegen: An open large language model for code with multi-turn program synthesis.
\newblock \emph{The Eleventh International Conference on Learning Representations}, 2022.

\bibitem[Olausson et~al.(2023)Olausson, Inala, Wang, Gao, and Solar-Lezama]{olausson2023self}
Theo~X Olausson, Jeevana~Priya Inala, Chenglong Wang, Jianfeng Gao, and Armando Solar-Lezama.
\newblock Is self-repair a silver bullet for code generation?
\newblock In \emph{The Twelfth International Conference on Learning Representations}, 2023.

\bibitem[{OpenAI}(2022)]{OpenAI2022}
{OpenAI}.
\newblock {ChatGPT}, 2022.
\newblock URL \url{https://openai.com/blog/chatgpt/}.

\bibitem[{OpenAI}(2024)]{OpenAI2024GPT4o}
{OpenAI}.
\newblock {GPT-4o-mini}, 2024.
\newblock URL \url{https://openai.com/index/gpt-4o-mini-advancing-cost-efficient-intelligence/}.

\bibitem[OpenAI \& Josh~Achiam(2024)OpenAI and Josh~Achiam]{OpenAI2023}
OpenAI and et~al. Josh~Achiam.
\newblock Gpt-4 technical report, 2024.
\newblock URL \url{https://arxiv.org/abs/2303.08774}.

\bibitem[Reynolds \& McDonell(2021)Reynolds and McDonell]{reynolds2021prompt}
Laria Reynolds and Kyle McDonell.
\newblock Prompt programming for large language models: Beyond the few-shot paradigm.
\newblock In \emph{Extended abstracts of the 2021 CHI conference on human factors in computing systems}, pp.\  1--7, 2021.

\bibitem[Rose et~al.(2010)Rose, Engel, Cramer, and Cowley]{rose2010automatic}
Stuart Rose, Dave Engel, Nick Cramer, and Wendy Cowley.
\newblock Automatic keyword extraction from individual documents.
\newblock \emph{Text mining: applications and theory}, pp.\  1--20, 2010.

\bibitem[Sparck~Jones(1972)]{sparck1972statistical}
Karen Sparck~Jones.
\newblock A statistical interpretation of term specificity and its application in retrieval.
\newblock \emph{Journal of documentation}, 28\penalty0 (1):\penalty0 11--21, 1972.

\bibitem[Sprague et~al.(2024)Sprague, Yin, Rodriguez, Jiang, Wadhwa, Singhal, Zhao, Ye, Mahowald, and Durrett]{sprague2024cot}
Zayne Sprague, Fangcong Yin, Juan~Diego Rodriguez, Dongwei Jiang, Manya Wadhwa, Prasann Singhal, Xinyu Zhao, Xi~Ye, Kyle Mahowald, and Greg Durrett.
\newblock To cot or not to cot? chain-of-thought helps mainly on math and symbolic reasoning.
\newblock \emph{arXiv preprint arXiv:2409.12183}, 2024.

\bibitem[Sun et~al.(2020)Sun, Qiu, Zheng, Wang, and Zhang]{sun2020sifrank}
Yi~Sun, Hangping Qiu, Yu~Zheng, Zhongwei Wang, and Chaoran Zhang.
\newblock Sifrank: a new baseline for unsupervised keyphrase extraction based on pre-trained language model.
\newblock \emph{IEEE Access}, 8:\penalty0 10896--10906, 2020.

\bibitem[Tang et~al.(2023)Tang, Wang, Chen, Wang, Liu, Chen, and Lin]{tang2023towards}
Ziyi Tang, Ruilin Wang, Weixing Chen, Keze Wang, Yang Liu, Tianshui Chen, and Liang Lin.
\newblock Towards causalgpt: A multi-agent approach for faithful knowledge reasoning via promoting causal consistency in llms.
\newblock \emph{arXiv preprint arXiv:2308.11914}, 2023.

\bibitem[Tian \& Chen(2023)Tian and Chen]{tian2023test}
Zhao Tian and Junjie Chen.
\newblock Test-case-driven programming understanding in large language models for better code generation.
\newblock \emph{arXiv preprint arXiv:2309.16120}, 2023.

\bibitem[Vaithilingam et~al.(2022)Vaithilingam, Zhang, and Glassman]{vaithilingam2022expectation}
Priyan Vaithilingam, Tianyi Zhang, and Elena~L Glassman.
\newblock Expectation vs. experience: Evaluating the usability of code generation tools powered by large language models.
\newblock In \emph{Chi conference on human factors in computing systems extended abstracts}, pp.\  1--7, 2022.

\bibitem[Wan \& Xiao(2008)Wan and Xiao]{wan2008single}
Xiaojun Wan and Jianguo Xiao.
\newblock Single document keyphrase extraction using neighborhood knowledge.
\newblock In \emph{AAAI}, volume~8, pp.\  855--860, 2008.

\bibitem[Wang et~al.(2023)Wang, Wei, Schuurmans, Le, Chi, Narang, Chowdhery, and Zhou]{wang2022self}
Xuezhi Wang, Jason Wei, Dale Schuurmans, Quoc~V. Le, Ed~H. Chi, Sharan Narang, Aakanksha Chowdhery, and Denny Zhou.
\newblock Self-consistency improves chain of thought reasoning in language models.
\newblock \emph{The Eleventh International Conference on Learning Representations, {ICLR} 2023, Kigali, Rwanda, May 1-5, 2023}, 2023.
\newblock URL \url{https://openreview.net/forum?id=1PL1NIMMrw}.

\bibitem[Wang et~al.(2021)Wang, Wang, Joty, and Hoi]{wang2021codet5}
Yue Wang, Weishi Wang, Shafiq Joty, and Steven~CH Hoi.
\newblock Codet5: Identifier-aware unified pre-trained encoder-decoder models for code understanding and generation.
\newblock In \emph{Proceedings of the 2021 Conference on Empirical Methods in Natural Language Processing}, pp.\  8696--8708, 2021.

\bibitem[Wang et~al.(2024)Wang, Liu, Lin, Li, Ma, and Liang]{wang2024rat}
Zihao Wang, Anji Liu, Haowei Lin, Jiaqi Li, Xiaojian Ma, and Yitao Liang.
\newblock Rat: Retrieval augmented thoughts elicit context-aware reasoning in long-horizon generation.
\newblock \emph{arXiv preprint arXiv:2403.05313}, 2024.

\bibitem[Wei et~al.(2022)Wei, Wang, Schuurmans, Bosma, Xia, Chi, Le, Zhou, et~al.]{wei2022chain}
Jason Wei, Xuezhi Wang, Dale Schuurmans, Maarten Bosma, Fei Xia, Ed~Chi, Quoc~V Le, Denny Zhou, et~al.
\newblock Chain-of-thought prompting elicits reasoning in large language models.
\newblock \emph{Advances in neural information processing systems}, 35:\penalty0 24824--24837, 2022.

\bibitem[Wiseman \& Rush(2016)Wiseman and Rush]{wiseman2016sequence}
Sam Wiseman and Alexander~M Rush.
\newblock Sequence-to-sequence learning as beam-search optimization.
\newblock In \emph{Proceedings of the 2016 Conference on Empirical Methods in Natural Language Processing}, pp.\  1296--1306, 2016.

\bibitem[Wu et~al.(2024)Wu, Hu, Shi, Dziri, Suhr, Ammanabrolu, Smith, Ostendorf, and Hajishirzi]{wu2024fine}
Zeqiu Wu, Yushi Hu, Weijia Shi, Nouha Dziri, Alane Suhr, Prithviraj Ammanabrolu, Noah~A Smith, Mari Ostendorf, and Hannaneh Hajishirzi.
\newblock Fine-grained human feedback gives better rewards for language model training.
\newblock \emph{Advances in Neural Information Processing Systems}, 36, 2024.

\bibitem[Xu et~al.(2022)Xu, Vasilescu, and Neubig]{xu2022ide}
Frank~F Xu, Bogdan Vasilescu, and Graham Neubig.
\newblock In-ide code generation from natural language: Promise and challenges.
\newblock \emph{ACM Transactions on Software Engineering and Methodology (TOSEM)}, 31\penalty0 (2):\penalty0 1--47, 2022.

\bibitem[Yang et~al.(2023)Yang, Zhang, Chen, Petzold, Wang, and Cheng]{yang2023zero}
Xianjun Yang, Kexun Zhang, Haifeng Chen, Linda Petzold, William~Yang Wang, and Wei Cheng.
\newblock Zero-shot detection of machine-generated codes.
\newblock \emph{arXiv preprint arXiv:2310.05103}, 2023.

\bibitem[Yao et~al.(2024)Yao, Yu, Zhao, Shafran, Griffiths, Cao, and Narasimhan]{yao2024tree}
Shunyu Yao, Dian Yu, Jeffrey Zhao, Izhak Shafran, Tom Griffiths, Yuan Cao, and Karthik Narasimhan.
\newblock Tree of thoughts: Deliberate problem solving with large language models.
\newblock \emph{Advances in Neural Information Processing Systems}, 36, 2024.

\bibitem[Yin \& Neubig(2017)Yin and Neubig]{yin2017syntactic}
Pengcheng Yin and Graham Neubig.
\newblock A syntactic neural model for general-purpose code generation.
\newblock In \emph{Proceedings of the 55th Annual Meeting of the Association for Computational Linguistics (Volume 1: Long Papers)}, pp.\  440--450, 2017.

\bibitem[Yu et~al.(2024)Yu, Jiang, Luo, Wu, Lin, Li, Yang, Huang, and Qiu]{yu2024mitigate}
Yijiong Yu, Huiqiang Jiang, Xufang Luo, Qianhui Wu, Chin-Yew Lin, Dongsheng Li, Yuqing Yang, Yongfeng Huang, and Lili Qiu.
\newblock Mitigate position bias in large language models via scaling a single dimension.
\newblock \emph{arXiv preprint arXiv:2406.02536}, 2024.

\bibitem[Zhang et~al.(2024)Zhang, Wang, Ren, Jiang, Wang, and Liu]{zhang2024ratt}
Jinghan Zhang, Xiting Wang, Weijieying Ren, Lu~Jiang, Dongjie Wang, and Kunpeng Liu.
\newblock Ratt: Athought structure for coherent and correct llmreasoning.
\newblock \emph{arXiv preprint arXiv:2406.02746}, 2024.

\bibitem[Zhong et~al.(2024{\natexlab{a}})Zhong, Wang, and Shang]{zhong2024debug}
Li~Zhong, Zilong Wang, and Jingbo Shang.
\newblock Debug like a human: A large language model debugger via verifying runtime execution step by step.
\newblock In \emph{Findings of the Association for Computational Linguistics ACL 2024}, pp.\  851--870, 2024{\natexlab{a}}.

\bibitem[Zhong et~al.(2024{\natexlab{b}})Zhong, Wang, and Shang]{zhong2024ldb}
Li~Zhong, Zilong Wang, and Jingbo Shang.
\newblock Ldb: A large language model debugger via verifying runtime execution step-by-step.
\newblock \emph{arXiv preprint arXiv:2402.16906}, 2024{\natexlab{b}}.

\bibitem[Zhu et~al.(2024)Zhu, Guo, Shao, Yang, Wang, Xu, Wu, Li, Gao, Ma, et~al.]{zhu2024deepseek}
Qihao Zhu, Daya Guo, Zhihong Shao, Dejian Yang, Peiyi Wang, Runxin Xu, Y~Wu, Yukun Li, Huazuo Gao, Shirong Ma, et~al.
\newblock Deepseek-coder-v2: Breaking the barrier of closed-source models in code intelligence.
\newblock \emph{arXiv preprint arXiv:2406.11931}, 2024.

\end{thebibliography}
\bibliographystyle{iclr2024_conference}
\clearpage

\appendix

\section{Aglorithm of KeyRank}
\label{appendix:alg_keyrank}
\begin{algorithm}
\small
\caption{KeyRank Procedure}
\label{alg_keyrank}
\begin{algorithmic}[1]
\Require Keyword Set $K_x$, Problem $\mathbf{P}$, Corpus $C$
\Ensure Ranked Keywords $K_y$

    \State $K_g \gets \emptyset$, $K_a \gets \emptyset$, $K_y \gets \emptyset$
    \State $\mathbf{f} \gets \textsc{ExtractFunctionName}(\mathbf{P})$
    
    \For{each $k$ in $K_x$}
        \If{$k = \mathbf{f}$}
            \State $K_g \gets K_g \cup \{(k, -1)\}$
        \ElsIf{$k \in \mathbf{P}$}
            \State $K_g \gets K_g \cup \{(k, \textsc{TF-IDF}(k, \mathbf{P}, C))\}$
        \Else
            \State $K_a \gets K_a \cup \{k\}$
        \EndIf
    \EndFor
    
    \State $K_g \gets \textsc{SortDescending}(K_g)$
    \State $K_y \gets K_a \cup K_g$
    
    \State \Return $K_y$
\end{algorithmic}
\end{algorithm}

First, we initialize the \textit{General Keywords}, \textit{Abstract Keywords}, and output as $K_g,K_a,K_y$, respectively. \textsc{ExtractFunctionName} extracts the method name if provided in the problem description. Otherwise, it returns a null value. Then, keywords are classified and scored. They can be divided into three classes: \textit{Abstract Keywords}, \textit{General Keywords}, and \textit{Function Keyword}. Abstract keywords do not appear in any input; they are abstract terms summarized from multiple concepts and stored in $K_a$. General keywords denote items in the problem description. We calculate their importance using TF-IDF based on a code-related corpus. General keywords and their scores are stored in $K_g$. Function keyword refers to the method name for solving the problem. Its explanation provides a coarse-grained description of the problem requirements. We assign a score of -1 to the function keyword, and also store them in $K_g$. Finally, \textsc{SortDescending} sorts the keywords in $K_g$ based on their scores. The keywords are combined in the order of abstract, general, and function keywords, and are then returned as the Ranked Keywords. 

\section{Studied LLMs}
\label{appendix:LLMs}
\begin{itemize}[leftmargin=*,noitemsep]
\item[$\bullet$] \textbf{Llama-3.1-70B}~\citep{dubey2024llama3herdmodels} is an open-sourced, decoder-only language model, pre-trained on 15t tokens from public sources. In our experiments, we use the Llama-3.1-70B-Instruct version.

\item[$\bullet$] \textbf{Mixtral-8$\times$22B}~\citep{jiang2024mixtralexperts}  is an open-source, sparse Mixture-of-Experts (MOE) model with 141B total parameters, utilizing 39B active parameters. We use the Mixtral-8$\times$22B-Instruct-v0.1 version.

\item[$\bullet$] \textbf{DeepSeek-Coder-V2-Instruct-0724}~\citep{zhu2024deepseek}, developed by DeepSeek-AI, is an open-source MoE code language model pre-trained on 10.2T tokens. The instruction-tuned version is further trained on 11B tokens. 

\item[$\bullet$] \textbf{GPT-3.5-turbo-0125}~\citep{OpenAI2022} is a close-sourced LLM from OpenAI, building on GPT-3 with optimizations for more efficient text generation. 
\item[$\bullet$] \textbf{GPT-4o-mini}~\citep{OpenAI2024GPT4o} is a smaller, cost-effective\footnote{GPT-4 is not selected due to the high experimental cost required.} variant of GPT-4~\citep{OpenAI2023}, offering strong performance across various tasks.
\end{itemize}

\section{Benchmark Details}
\begin{table}[H]
\centering
\resizebox{0.95\linewidth}{!}{
\begin{tabular}{@{}lccccccc@{}}
\toprule
Benchmark & Humaneval & Humaneval+ & MBPP & MBPP+ & \begin{tabular}[c]{@{}c@{}}APPS\\ Introductory\end{tabular} & \begin{tabular}[c]{@{}c@{}}APPS\\ Interview\end{tabular} & \begin{tabular}[c]{@{}c@{}}APPS\\ Competition\end{tabular} \\ \midrule
Problem & 164 & 164 & 399 & 399 & 60 & 180 & 60 \\
\#Avg Tests & 9.6 & 764.1 & 3.1 & 105.4 & 15.1 & 25.7 & 17.3 \\
\#Avg Tokens & 67.7 & 67.7 & 26.1 & 26.1 & 257.3 & 319.8 & 377.4 \\ \bottomrule
\end{tabular}
}
\caption{Statistics of benchmarks: the total number of problems in each benchmark (Problems), the average number of hidden test cases per problem (\#Avg Tests), and the average number of space-separated tokens of the problem (\#Avg Tokens).}
\label{tab_benchmark}
\end{table}
\label{appendix:benchmark}

We use three widely-used benchmarks, i.e., HumanEval(+), MBPP(+), and APPS, for evaluation. 
Table \ref{tab_benchmark} presents their key statistics.\\
(1) \textbf{HumanEval}~\citep{chen2021evaluating} consists of 164 hand-written programming problems, each including a method signature, docstring, body, and unit tests. We use both HumanEval and its extended version, HumanEval+\citep{liu2024your}, which enhances the original with 80$\times$ additional test samples to address test case insufficiency~\citep{liu2024your}.\\
(2) \textbf{MBPP}~\citep{austin2021program} contains crowd-sourced Python programming problems. Our study uses the versions proposed by~\citep{liu2024your}, including MBPP and MBPP+. Each of them contain 399 tasks, and the latter adds 35$\times$ test samples. \\
(3) \textbf{APPS}~\citep{hendrycks2measuring} includes 10,000 coding problems from open-access websites, split equally into training and test sets. It includes two problem formats: call-based format (input via function parameters) and standard input format (using stdin/stdout). Problems are categorized into introductory, interview, and competition levels. There are three different difficulty levels of problems in APPS, i.e., introductory, interview and competition. Each of them has 1000, 3000, and 1000 tasks, respectively. Considering the cost of evaluating the entire APPS test set and following prior work~\citep{olausson2023self,huang2024knowledge,lecodechain,yang2023zero}, we randomly select problems in accordance with the frequency distribution of these difficulty levels and sample 60, 180, 60 problems at the introductory, interview, and competition levels, respectively.

\section{Implementation Details}
\label{appendix:details}
\textbf{Demonstration selection strategy.} Specifically, for HumanEval, we select the first two problems as demonstrations. For MBPP, we choose the first problem. For APPS, considering the model's input length limitation and to avoid randomness, we select the two shortest problems from the first five problems in the training set. The reason for this differentiated strategy is that HumanEval and APPS problems are more complex, requiring more examples, while MBPP problems are relatively simple in form, and one example is enough. 

\textbf{Keywords and explanations involved in demonstrations.} The prompt for KeyExtract \& Explain uses several demonstrations to guide LLMs to produce keywords and their explanations. To ensure the quality of each demonstration, we first employ Claude-3.5-Sonnet, an LLM separate from our target LLMs, to generate multiple sets of keywords and explanations for each demonstration. The generated contents are then manually reviewed, and the most accurate set for each demonstration is selected and used in the prompt. This can mitigate the potential bias in human-generated explanations. Additionally, for HumanEval(+) and MBPP(+) datasets, which provide function names, the first two authors discuss and write the explanation for the function name in each demonstration.

\section{Additional experiments}
\subsection{Influence of Keyword Combination Orders}
\label{append:orders}

\begin{table}[t]
\centering
\resizebox{0.75\linewidth}{!}{
\begin{tabular}{@{}c|c|cc@{}}
\toprule
Model & Combination Order & HumanEval & HumanEval+ \\ \midrule
\multirow{4}{*}{Llama-3.1-70B-Instruct} & Func\_Abs\_Gen & 83.5 & 78.7 \\
 & Gen\_Func\_Abs & 84.1 & 78.7 \\
 & Gen\_Abs\_Func & 84.1 & 78.7 \\
 & \SEK & \textbf{84.8} & \textbf{79.3} \\ \midrule
\multirow{4}{*}{Mixtral-8×22B-Instruct-v0.1} & Func\_Abs\_Gen & 78.0 & 72.0 \\
 & Gen\_Func\_Abs & 78.0 & 72.0 \\
 & Gen\_Abs\_Func & 76.8 & 71.3 \\
 & \SEK & \textbf{81.1} & \textbf{75.6} \\ \bottomrule
\end{tabular}
}
\caption{The experiments of different combination orders on Humaneval(+) with two LLMs.}

\label{tab:combation_order}
\end{table}

In KeyRank, we combine different types of keywords based on the order of \textit{abstract $\rightarrow$ general $\rightarrow$ function}.
We investigate the influence of keyword combination orders by comparing the order used by \SEK with three alternative ordering strategies using two LLMs, i.e., Llama-3.1-70B-Instruct and Mixtral-8×22B-Instruct-v0.1. 
Table~\ref{tab:combation_order} presents the experimental results, where the abbreviations Abs, Gen, and Func denote \textit{abstract keywords}, \textit{general keywords}, and \textit{function keywords}, respectively.
The results reveal performance variations across different keyword combination orders, indicating that the order of different keyword types impacts LLMs' comprehension of coding problems. The combination order used by \SEK consistently yields optimal performance, suggesting its rationality.

\subsection{Influence of Guidelines}
\begin{table}[t]
\centering
\resizebox{0.75\linewidth}{!}{
\begin{tabular}{@{}c|c|cc@{}}
\toprule
Model & Ablations & HumanEval & HumanEval+ \\ \midrule
\multirow{6}{*}{Llama-3.1-70B-Instruct} & w/o Guideline(1) & \textbf{85.4} & 78.7 \\
 & w/o Guideline(2) & 82.3 & 75.6 \\
 & w/o Guideline(3) & 81.7 & 76.8 \\
 & w/o Guideline(4) & 81.1 & 76.2 \\
 & w/o Guideline(5) & 83.5 & 77.4 \\
 & ALL Guidelines & 84.8 & \textbf{79.3} \\ \midrule
\multirow{6}{*}{Mixtral-8×22B-Instruct-v0.1} & w/o Guideline(1) & 76.8 & 72 \\
 & w/o Guideline(2) & 77.4 & 72.6 \\
 & w/o Guideline(3) & 79.3 & 73.8 \\
 & w/o Guideline(4) & 75.0 & 70.1 \\
 & w/o Guideline(5) & 76.8 & 73.2 \\
 & ALL Guidelines & \textbf{81.1} & \textbf{75.6} \\ \bottomrule
\end{tabular}
}
\caption{Ablation experiments on removing one guideline at a time from Keyword Prompt on HumanEval(+) with two LLMs.}
\label{tab:guideline_ablation}
\end{table}

\label{append:guide}
In Section~\ref{sec:discussion}, we investigate the effectiveness of the guidelines in the KeyExtract \& Explain prompt as a whole.
This section further investigates the impact of each guideline by removing it from the prompt and re-evaluate the performance of \SEK with two LLMs, i.e., Llama-3.1-70B-Instruct and Mixtral-8×22B-Instruct-v0.1 on HumanEval(+).
Table~\ref{tab:guideline_ablation} presents the experimental results, where the performance of the two LLMs decreases in almost all cases, indicating the contribution of each guideline to the effectiveness of SEK.

\subsection{More Experiments on APPS}
\label{append:random_apps}
\begin{table}[t]
\centering
\resizebox{\linewidth}{!}{
\begin{tabular}{@{}cccccc@{}}
\toprule
Model & Method & Introductory(A) & Introductory(B) & Introductory(C) & Average \\ \midrule
\multirow{5}{*}{Llama-3.1-70B-Instruct} & Default & 51.6 & 45.0 & 46.6 & 47.7 \\
 & Beam Search & 55.0 & 45.0 & 45.0 & 48.3 \\
 & CoT & 41.6 & 46.6 & 45.0 & 44.4 \\
 & SelfEvolve & 45.0 & 53.3 & 46.6 & 48.3 \\ \cmidrule(l){2-6} 
 & \textbf{SEK} & \textbf{58.3} & \textbf{56.6} & \textbf{50.0} & \textbf{55.0} \\ \midrule
\multirow{4}{*}{\begin{tabular}[c]{@{}c@{}}GPT-3.5-turbo\\ (API)\end{tabular}} & Default & 45.0 & 51.6 & 43.3 & 46.6 \\
 & CoT & 48.3 & 53.3 & 46.6 & 49.4 \\
 & SelfEvolve & 46.6 & 48.3 & 45.0 & 46.6 \\ \cmidrule(l){2-6} 
 & \textbf{SEK} & \textbf{48.3} & \textbf{53.3} & \textbf{50.0} & \textbf{50.5} \\ \bottomrule
\end{tabular}
}
\caption{The Pass@1 (\%) results of \SEK and baseline methods on differently sampled APPS-Introductory sets.}
\label{tab:different_apps}
\end{table}

In the main experiment, we randomly sample problems from the APPS test set for evaluation due to limited resources. The performance of LLMs on APPS may be affected by the randomness of the selected samples. To mitigate this variability, we conduct additional experiments by randomly selecting three new subsets of problems at the introductory level from the APPS test set and using two LLMs for evaluation, i.e., Llama-3.1-70B-instruct and GPT-3.5-Turbo. The number of sampled tasks is fixed at 60, consistent with the main experiment. For reproducibility, the selected tasks are provided in Table~\ref{tab:introductory_ids}.
As shown in Table~\ref{tab:different_apps}, \SEK achieve optimal performance across different subsets. For instance, considering Llama-3.1-70B-Instruct,  \SEK outperforms the Default, Beam Search, and CoT baselines by an average of 7.3\%, 6.7\%, and 10.6\%, respectively. This corroborates the credibility of our conclusions.

\begin{table}[t]
\centering
\resizebox{0.85\linewidth}{!}{
\begin{tabular}{@{}p{3cm}|p{10cm}@{}}
\toprule
\multicolumn{1}{c|}{Datasets} & \multicolumn{1}{c}{Tasks} \\ \midrule
Introductory(A) & 4029, 4032, 4050, 4054, 4060, 4099, 4116, 4131, 4132, 4148, 4157, 4166, 4180, 4206, 4211, 4232, 4251, 4256, 4283, 4289, 4317, 4320, 4323, 4332, 4343, 4356, 4417, 4451, 4469, 4471, 4527, 4538, 4541, 4542, 4546, 4599, 4625, 4640, 4676, 4680, 4704, 4721, 4748, 4774, 4780, 4781, 4787, 4800, 4806, 4826, 4837, 4864, 4868, 4878, 4888, 4896, 4924, 4926, 4930, 4943 \\ \midrule
Introductory(B) & 4021, 4046, 4051, 4073, 4115, 4127, 4138, 4140, 4156, 4163, 4201, 4225, 4230, 4233, 4236, 4263, 4270, 4294, 4295, 4347, 4358, 4375, 4376, 4407, 4424, 4446, 4453, 4454, 4460, 4469, 4478, 4486, 4489, 4528, 4547, 4570, 4580, 4596, 4638, 4644, 4656, 4678, 4692, 4695, 4726, 4730, 4735, 4740, 4780, 4803, 4842, 4869, 4871, 4890, 4905, 4918, 4932, 4947, 4970, 4975 \\ \midrule
Introductory(C) & 4018, 4028, 4042, 4085, 4102, 4134, 4146, 4172, 4201, 4203, 4257, 4274, 4281, 4309, 4314, 4324, 4353, 4356, 4387, 4418, 4447, 4461, 4465, 4473, 4474, 4486, 4499, 4506, 4510, 4528, 4560, 4592, 4604, 4617, 4627, 4701, 4704, 4710, 4712, 4713, 4716, 4719, 4723, 4732, 4763, 4772, 4801, 4812, 4867, 4879, 4893, 4898, 4907, 4924, 4929, 4944, 4957, 4973, 4974, 4975 \\ \bottomrule
\end{tabular}
}
\caption{The tasks in different sampling Introductory sets.
}
\label{tab:introductory_ids}
\end{table}

\section{Selected APPS tasks}
\label{append:selected_apps}
For reproducibility, we provide the complete list of selected tasks of APPS in Table~\ref{tab:different_difficulty}.
\begin{table}[t]
\centering
\resizebox{0.85\linewidth}{!}{
\begin{tabular}{@{}p{3cm}|p{10cm}@{}}
\toprule
Difficulty & Tasks \\ \midrule
Introductory & 4007, 4032, 4049, 4050, 4054, 4060, 4114, 4116, 4132, 4148, 4157, 4166, 4180, 4211, 4215, 4232, 4251, 4283, 4289, 4317, 4323, 4332, 4343, 4356, 4372, 4417, 4439, 4451, 4469, 4527, 4540, 4541, 4546, 4549, 4582, 4585, 4599, 4625, 4631, 4640, 4676, 4678, 4704, 4721, 4774, 4781, 4787, 4800, 4806, 4826, 4837, 4861, 4864, 4868, 4878, 4888, 4924, 4926, 4929, 4930 \\ \midrule
Interview & 6, 10, 35, 44, 56, 76, 82, 95, 105, 106, 115, 133, 135, 178, 188, 198, 210, 213, 231, 240, 248, 278, 300, 305, 319, 342, 357, 372, 377, 379, 420, 457, 460, 483, 484, 489, 546, 553, 566, 567, 584, 634, 664, 669, 675, 686, 696, 701, 734, 785, 817, 855, 861, 876, 903, 909, 914, 932, 973, 989, 993, 994, 1017, 1020, 1025, 1033, 1039, 1053, 1069, 1101, 1122, 1132, 1140, 1144, 1158, 1166, 1167, 1224, 1226, 1232, 1280, 1313, 1346, 1351, 1361, 1373, 1375, 1391, 1394, 1406, 1409, 1432, 1458, 1459, 1478, 1487, 1491, 1508, 1520, 1527, 1534, 1540, 1557, 1563, 1565, 1590, 1635, 1640, 1715, 1720, 1733, 1749, 1761, 1768, 1775, 1813, 1823, 1833, 1838, 1864, 1881, 1885, 1955, 1976, 1982, 1989, 2003, 2006, 2011, 2015, 2048, 2053, 2062, 2077, 2097, 2101, 2145, 2177, 2192, 2209, 2273, 2293, 2317, 2361, 2406, 2443, 2492, 2494, 2495, 2502, 2513, 2514, 2533, 2542, 2546, 2552, 2554, 2609, 2615, 2641, 2642, 2655, 2657, 2684, 2707, 2725, 2726, 2728, 2729, 2762, 2767, 2776, 2784, 2788, 2815, 2850, 2874, 2914, 2982, 2999 \\ \midrule
Competition & 3009, 3024, 3031, 3071, 3097, 3131, 3138, 3171, 3188, 3204, 3206, 3210, 3211, 3252, 3262, 3263, 3298, 3301, 3313, 3319, 3326, 3372, 3379, 3445, 3456, 3479, 3481, 3501, 3517, 3535, 3573, 3579, 3618, 3629, 3654, 3680, 3684, 3690, 3713, 3721, 3727, 3731, 3733, 3745, 3762, 3775, 3786, 3788, 3802, 3803, 3843, 3863, 3882, 3886, 3893, 3901, 3943, 3945, 3948, 3972 \\ \bottomrule
\end{tabular}
}
\caption{The tasks in different difficulty levels of APPS.
}
\label{tab:different_difficulty}
\end{table}

\section{Attention Analysis}
\label{appendix:attention}
We aim to explain \SEK from the perspective of attention distribution.
We use BertViz\footnote{https://github.com/jessevig/bertviz} to present explainability visualizations. Due to limited computational resources, we select a short problem and remove its test cases. Specifically, the problem description is ``Write a function to find the nth nonagonal number.'' and we select a keyword with its explanation ``[nonagonal]: A nine-sided polygon. Nonagonal numbers represent the count of dots forming nonagons of increasing size''. We select Mixtral-8×22B-Instruct-v0.1 as the base model and extract the attention from its last layer for analysis.

\begin{figure}[t]
\centering
\begin{minipage}{0.35\linewidth}
    \label{fig:attention1}
    \centering
    \resizebox{0.8\linewidth}{!}{
        \includegraphics[angle=90]{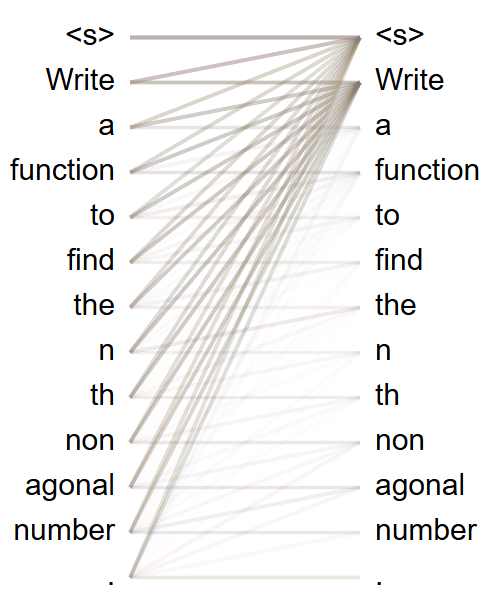}
    }
    \caption{Attention visualization with Default.}
\end{minipage}%
\hspace{0.05\linewidth}%
\begin{minipage}{0.35\linewidth}
    \centering
    \label{fig:attention2}

    \resizebox{0.8\linewidth}{!}{
        \includegraphics[angle=90]{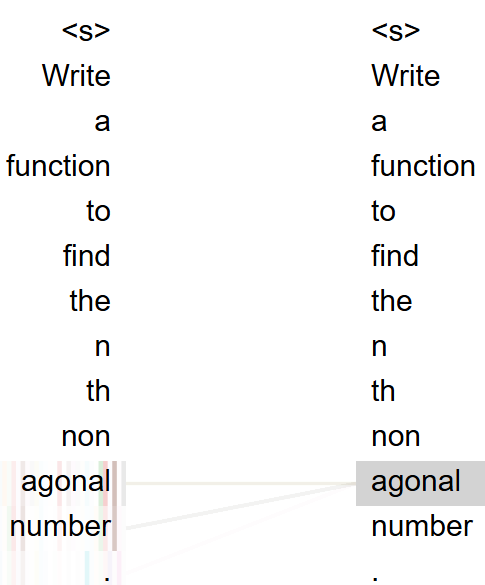}
    }
    \caption{Attention visualization for a part of the keyword "nonagonal" with Default.}
\end{minipage}
\end{figure}

\begin{figure}[t]
\centering
\resizebox{\linewidth}{!}{
\begin{minipage}[t]{\linewidth}
    \centering
    \includegraphics[width=0.2\textwidth,angle=90]{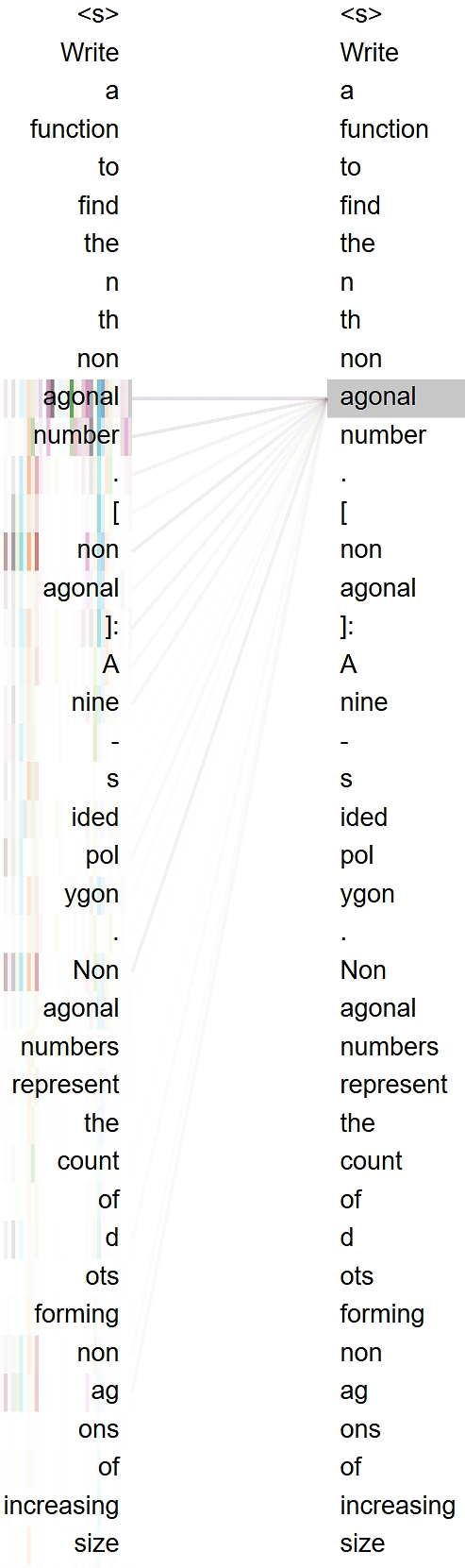}
    \caption{Attention visualization for a part of the keyword "nonagonal" with  \SEK.}
\end{minipage}
}
\label{fig:attention3}
\end{figure}

\begin{figure}[H]
\label{fig:final}
\centering
\resizebox{\linewidth}{!}{
\begin{minipage}[t]{\linewidth}
    \centering
    \includegraphics[width=0.25\textwidth,angle=90]{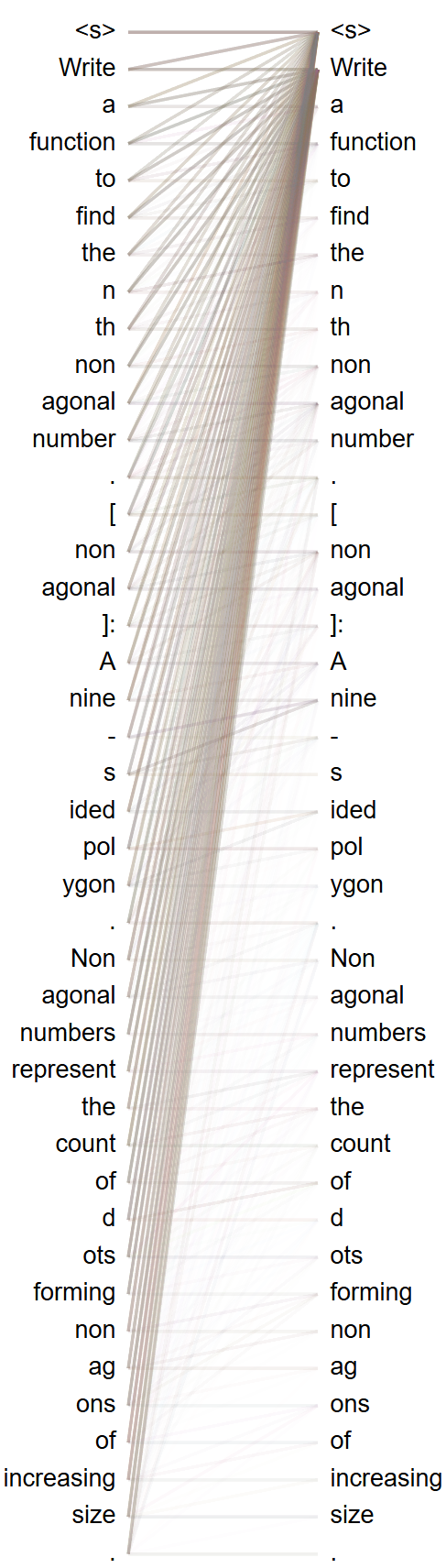}
    \caption{Final layer of the attention visualization with \SEK}
\end{minipage}
}

\end{figure}

The key to this problem lies in understanding ``nonagonal''. With Default, Figure \ref{fig:attention1} shows the overall attention distribution for the problem, while Figure \ref{fig:attention2} displays the attention distribution for a part of the keyword ``nonagonal". It can be observed that most of the attention is allocated to the beginning words, with the keyword ``nonagonal" receiving relatively less attention. This may lead to insufficient focus on the core concept of the problem when generating code. In contrast, with \SEK, Figure~\ref{fig:final} presents the overall attention distribution of the LLM with \SEK and Figure \ref{fig:attention3} shows the attention distribution for ``nonagonal".  It can be seen that the model allocates additional attention to the added keywords and explanations, encouraging the model to focus more on the core concepts of the problem. With \SEK, the LLM further distributes attention to the added keywords and explanations, which can enhance its understanding of the key concepts in the problem.


\section{Prompt For Self-Evolve}
\label{appendix:prompt_evolve}

\begin{lstlisting}[caption=The first prompt of Self-Evolve]
``` 
{Problem description} 
```  
For the above question, could you briefly teach me how to solve it step by step in natural language? Don’t write the code in this step.
\end{lstlisting}
\begin{lstlisting}[caption=The Second prompt of Self-Evolve]
Based on the above idea, help me complete the function.
Be attention, you should only output the codes without any explanation and natural language. Wrap your code with "```"
\end{lstlisting}

\section{Case Study of the Difference between Beam Search and SEK}
\label{appendix:beam}
\begin{lstlisting}[caption=A problem from Humaneval (problem ID $131$)]
def digits(n):
    """Given a positive integer n, return the product of the odd digits.
    Return 0 if all digits are even.
    For example:
    digits(1)  == 1
    digits(4)  == 0
    digits(235) == 15
    """
\end{lstlisting}
\begin{lstlisting}[caption=The output of Default from Humaneval (problem ID $131$)]
def digits(n):
    """
    Given a positive integer n, return the product of the odd digits.
    Return 0 if all digits are even.

    Args:
        n (int): A positive integer.

    Returns:
        int: The product of the odd digits of n, or 0 if all digits are even.
    """
    product = 1
    for digit in str(n):
        digit = int(digit)
        if digit % 2 != 0:
            product *= digit
    return product if product != 1 else 0
\end{lstlisting}
\begin{lstlisting}[caption=The output of Beam Search from Humaneval (problem ID $131$)]
def digits(n):
    """Given a positive integer n, return the product of the odd digits.
    Return 0 if all digits are even.
    For example:
    digits(1)  == 1
    digits(4)  == 0
    digits(235) == 15
    """
    product = 1
    for digit in str(n):
        if int(digit) % 2 != 0:
            product *= int(digit)
    return product if product != 1 else 0
\end{lstlisting}
\begin{lstlisting}[caption=The output of SEK from Humaneval (problem ID $131$)]
[positive integer n]: The input parameter, a whole number greater than zero, which is the subject of the operation. The test cases demonstrate this concept with inputs 1, 4, and 235.
[product of the odd digits]: The result of multiplying all the odd digits together. In the test case digits(235), the product of the odd digits is 3 * 5 = 15.
[odd digits]: Refers to the individual numerical digits within the input integer n that are not divisible by 2, i.e., 1, 3, 5, 7, and 9. In the test case digits(235), the odd digits are 2, 3, and 5.

def digits(n):
    """
    Given a positive integer n, return the product of the odd digits.
    Return 0 if all digits are even.

    Args:
        n (int): A positive integer.

    Returns:
        int: The product of the odd digits, or 0 if all digits are even.
    """
    # Convert the integer to a string to easily iterate over each digit
    str_n = str(n)
    
    # Initialize the product to 1
    product = 1
    
    # Initialize a flag to track if any odd digits were found
    found_odd = False
    
    # Iterate over each character (digit) in the string
    for char in str_n:
        # Convert the character back to an integer
        digit = int(char)
        
        # Check if the digit is odd
        if digit % 2 != 0:
            # If the digit is odd, multiply it with the current product
            product *= digit
            # Set the flag to True
            found_odd = True
    
    # If no odd digits were found, return 0
    if not found_odd:
        return 0
    # Otherwise, return the product of the odd digits
    else:
        return product
\end{lstlisting}

Despite expanding the search space, Beam Search is still less effective than SEK due to its failure to deepen its understanding of the problem. To illustrate this, we use Problem 131 from Humaneval, generated by Llama 3.1-70B-Instruct. Although Beam Search and the Default implementations differ, neither approach fully comprehends the problem. Specifically, when handling \textit{odd digits}, both methods incorrectly return 0 when the product of the \textit{odd digits} is 1. In contrast, SEK not only identifies but also correctly interprets the concept of \textit{odd digits} in the problem description, allowing it to handle cases where the product of the odd digits equals 1 accurately. This demonstrates that SEK, by focusing on the underlying semantic understanding of key problem concepts, develops a deeper comprehension of the task, ultimately leading to the correct solution.

\section{More examples}
\label{append:case}

We extract an example from MBPP. The problem involves finding an element with a 1-based index in a given array. It can be observed that both Default and CoT make the same mistake: they perform unnecessary sorting operations on the array, resulting in the generation of incorrect code solutions. In contrast, \SEK does not conduct sorting. It accurately extracts the core concepts of the problem, i.e., \textit{1-based indexing} and \textit{kth element}. Explanations of these two key concepts do not mention any need for sorting, demonstrating a precise grasp of the problem's requirements. Consequently, \SEK successfully generated the correct code solution.

\begin{lstlisting}[caption=A problem from MBPP (problem ID $101$)]
Write a function to find the kth element in the given array using 1-based indexing.
assert kth_element([12,3,5,7,19], 2) == 3
\end{lstlisting}
\begin{lstlisting}[caption=The output of Default from MBPP (problem ID $101$)]
def kth_element(arr, k):
    """
    Write a function to find the kth element in the given array using 1-based indexing.
    assert kth_element([12,3,5,7,19], 2) == 3
    """
    if k < 1 or k > len(arr):
        raise ValueError("k is out of bounds for the given array")
    return sorted(arr)[k-1]
\end{lstlisting}
\begin{lstlisting}[caption=The output of CoT from MBPP (problem ID $101$)]
def kth_element(arr, k):
    """
    Find the kth element in the given array using 1-based indexing.
    """
    # Ensure the array is sorted
    sorted_arr = sorted(arr)
    # Return the kth element using 1-based indexing
    return sorted_arr[k - 1]
\end{lstlisting}
\begin{lstlisting}[caption=The keywords output of \SEK from MBPP (problem ID $101$)]
[1-based indexing]: A method of indexing where the first element of a sequence is assigned the index 1. This contrasts with 0-based indexing, where the first element is assigned the index 0. In the test case, the function is expected to use 1-based indexing, meaning the second element is accessed with the index 2, not 1.
[kth element]: The element at the k-th position in a sequence, where k is a specified index. In the context of the test case, kth_element([12, 3, 5, 7, 19], 2) == 3, the kth element refers to the second element in the array [12, 3, 5, 7, 19], which is 3.
\end{lstlisting}
\begin{lstlisting}[caption=The enriched problem of MBPP (problem ID $101$)]
Write a function to find the kth element in the given array using 1-based indexing.
assert kth_element([12,3,5,7,19], 2) == 3

Analyze the following key terms and their relationships within the problem context:
    [1-based indexing]: A method of indexing where the first element of a sequence is assigned the index 1. This contrasts with 0-based indexing, where the first element is assigned the index 0. In the test case, the function is expected to use 1-based indexing, meaning the second element is accessed with the index 2, not 1.
    [kth element]: The element at the k-th position in a sequence, where k is a specified index. In the context of the test case, kth_element([12, 3, 5, 7, 19], 2) == 3, the kth element refers to the second element in the array [12, 3, 5, 7, 19], which is 3.

assert kth_element([12,3,5,7,19], 2) == 3
\end{lstlisting}
\begin{lstlisting}[caption=The output of \SEK from MBPP (problem ID $101$)]
def kth_element(arr, k):
    """
    Find the kth element in the given array using 1-based indexing.
    Args:
    arr (list): The input array.
    k (int): The position of the element to find (1-based indexing).
    Returns:
    int: The kth element in the array.
    """
    if k < 1 or k > len(arr):
        raise ValueError("k is out of the range of the array indices")
    return arr[k - 1]
\end{lstlisting}

We also present an example from APPS-Introductory. This problem requires solving an optimization problem with the goal of minimizing the number of thieves while ensuring that each thief steals the same number of swords and accounting for all missing swords. This solution requires the use of the greatest common divisor (GCD). It can be seen that both CoT and Default make the same mistake: they use the difference between the maximum and minimum sword counts to calculate the number of swords each person takes. In contrast, only \SEK correctly utilizes the GCD, resulting in the generation of accurate code.

\begin{lstlisting}[caption=A problem from APPS-Introductory (problem ID $4372$)]
There were $n$ types of swords in the theater basement which had been used during the plays. Moreover there were exactly $x$ swords of each type. $y$ people have broken into the theater basement and each of them has taken exactly $z$ swords of some single type. Note that different people might have taken different types of swords. Note that the values $x, y$ and $z$ are unknown for you.

The next morning the director of the theater discovers the loss. He counts all swords — exactly $a_i$ swords of the $i$-th type are left untouched.

The director has no clue about the initial number of swords of each type in the basement, the number of people who have broken into the basement and how many swords each of them have taken.

For example, if $n=3$, $a = [3, 12, 6]$ then one of the possible situations is $x=12$, $y=5$ and $z=3$. Then the first three people took swords of the first type and the other two people took swords of the third type. Note that you don't know values $x, y$ and $z$ beforehand but know values of $n$ and $a$.

Thus he seeks for your help. Determine the minimum number of people $y$, which could have broken into the theater basement, and the number of swords $z$ each of them has taken.


-----Input-----

The first line of the input contains one integer $n$ $(2 \le n \le 2 \cdot 10^{5})$ — the number of types of swords.

The second line of the input contains the sequence $a_1, a_2, \dots, a_n$ $(0 \le a_i \le 10^{9})$, where $a_i$ equals to the number of swords of the $i$-th type, which have remained in the basement after the theft. It is guaranteed that there exists at least one such pair of indices $(j, k)$ that $a_j \neq a_k$.


-----Output-----

Print two integers $y$ and $z$ — the minimum number of people which could have broken into the basement and the number of swords each of them has taken.


-----Examples-----
Input
3
3 12 6

Output
5 3

Input
2
2 9

Output
1 7

Input
7
2 1000000000 4 6 8 4 2

Output
2999999987 2

Input
6
13 52 0 13 26 52

Output
12 13

-----Note-----
In the first example the minimum value of $y$ equals to $5$, i.e. the minimum number of people who could have broken into the basement, is $5$. Each of them has taken $3$ swords: three of them have taken $3$ swords of the first type, and two others have taken $3$ swords of the third type.
In the second example the minimum value of $y$ is $1$, i.e. the minimum number of people who could have broken into the basement, equals to $1$. He has taken $7$ swords of the first type.
\end{lstlisting}
\begin{lstlisting}[caption=The output of Default from APPS-Introductory (problem ID $4372$)]
Below is a Python script with a self-contained function that solves the problem and passes corresponding tests:
```python
import math

# Read the number of types of swords
n = int(input())
# Read the sequence of remaining swords
a = list(map(int, input().split()))

# Find the minimum and maximum values in the sequence
min_swords = min(a)
max_swords = max(a)

# Calculate the number of swords each person took
z = max_swords - min_swords

# Calculate the minimum number of people
y = 0
if z != 0:
    for swords in a:
        y += (swords - min_swords) // z

# Output the result
print(y, z)
```

This script reads the input, calculates the minimum and maximum values of the remaining swords, and then computes the number of people (`y`) and the number of swords each person took (`z`). It ensures that the output is the minimum number of people who could have broken into the basement and the number of swords each of them has taken.
\end{lstlisting}
\begin{lstlisting}[caption=The output of CoT from APPS-Introductory (problem ID $4372$)]
Below is a Python script with a self-contained function that solves the problem and passes corresponding tests:
```python
# Read the number of types of swords
n = int(input())
# Read the sequence of remaining swords
a = list(map(int, input().split()))

# Step 1: Find the minimum and maximum values in the array
min_swords = min(a)
max_swords = max(a)

# Step 2: Calculate the number of swords each person took
z = max_swords - min_swords

# Step 3: Calculate the minimum number of people
y = 0
for swords in a:
    if swords != max_swords:
        y += (max_swords - swords) // z

# Step 4: Print the result
print(y, z)
```

This script reads the input, calculates the minimum number of people (`y`) and the number of swords each person took (`z`), and prints the result. The logic is based on finding the difference between the maximum and minimum values in the array `a`, which represents the number of swords each person took (`z`), and then summing up the number of people required to achieve this difference for each type of sword.
\end{lstlisting}
\begin{lstlisting}[caption=The enriched problem of  APPS-Introductory (problem ID $4372$)]

There were $n$ types of swords in the theater basement which had been used during the plays. Moreover there were exactly $x$ swords of each type. $y$ people have broken into the theater basement and each of them has taken exactly $z$ swords of some single type. Note that different people might have taken different types of swords. Note that the values $x, y$ and $z$ are unknown for you.

The next morning the director of the theater discovers the loss. He counts all swords — exactly $a_i$ swords of the $i$-th type are left untouched.

The director has no clue about the initial number of swords of each type in the basement, the number of people who have broken into the basement and how many swords each of them have taken.

For example, if $n=3$, $a = [3, 12, 6]$ then one of the possible situations is $x=12$, $y=5$ and $z=3$. Then the first three people took swords of the first type and the other two people took swords of the third type. Note that you don't know values $x, y$ and $z$ beforehand but know values of $n$ and $a$.

Thus he seeks for your help. Determine the minimum number of people $y$, which could have broken into the theater basement, and the number of swords $z$ each of them has taken.


-----Input-----

The first line of the input contains one integer $n$ $(2 \le n \le 2 \cdot 10^{5})$ — the number of types of swords.

The second line of the input contains the sequence $a_1, a_2, \dots, a_n$ $(0 \le a_i \le 10^{9})$, where $a_i$ equals to the number of swords of the $i$-th type, which have remained in the basement after the theft. It is guaranteed that there exists at least one such pair of indices $(j, k)$ that $a_j \neq a_k$.


-----Output-----

Print two integers $y$ and $z$ — the minimum number of people which could have broken into the basement and the number of swords each of them has taken.


-----Examples-----
Input
3
3 12 6

Output
5 3

Input
2
2 9

Output
1 7

Input
7
2 1000000000 4 6 8 4 2

Output
2999999987 2

Input
6
13 52 0 13 26 52

Output
12 13

-----Note-----
In the first example the minimum value of $y$ equals to $5$, i.e. the minimum number of people who could have broken into the basement, is $5$. Each of them has taken $3$ swords: three of them have taken $3$ swords of the first type, and two others have taken $3$ swords of the third type.
In the second example the minimum value of $y$ is $1$, i.e. the minimum number of people who could have broken into the basement, equals to $1$. He has taken $7$ swords of the first type.

-----Keywords and Explanations-----
Analyze the following key terms and their relationships within the problem context:
[number of swords taken]: The quantity of swords, denoted as \( z \), that each person took. This value is derived from the GCD of the differences between the initial number of swords and the remaining number of swords for each type. In the first example, \( z = 3 \) is the result of calculating the GCD of the differences in the remaining swords.
[minimum number of people]: The smallest count of individuals, denoted as \( y \), who could have broken into the theater basement and taken swords. This value is determined by the differences in the number of swords left of each type, \( a_i \), and is calculated as the greatest common divisor (GCD) of these differences. In the first example, \( y = 5 \) is derived from the differences between the remaining swords of each type.
\end{lstlisting}
\begin{lstlisting}[caption=The output of \SEK from APPS-Introductory (problem ID $4372$)]
Below is a Python script that solves the problem described:

```python
import math
from functools import reduce

def gcd(a, b):
    while b:
        a, b = b, a % b
    return a

def find_gcd(list):
    x = reduce(gcd, list)
    return x

n = int(input())
a = list(map(int, input().split()))

# Calculate the differences between the maximum and each element in a
differences = [max(a) - ai for ai in a]

# Find the GCD of all differences to determine the number of swords each person took (z)
z = find_gcd(differences)

# Calculate the minimum number of people (y) by summing up the differences and dividing by z
y = sum(differences) // z

print(y, z)
```

This script reads the number of types of swords `n` and the list of remaining swords `a` from the input. It then calculates the differences between the maximum number of swords and each type's remaining swords. The greatest common divisor (GCD) of these differences is computed to find the number of swords each person took (`z`). Finally, it calculates the minimum number of people (`y`) by summing up the differences and dividing by `z`. The results are printed as `y` and `z`.
\end{lstlisting}

\begin{lstlisting}[caption=A problem from MBPP (problem ID $594$)]
Write a function to find the difference of the first even and first odd number of a given list.
assert diff_even_odd([1,3,5,7,4,1,6,8])==3
\end{lstlisting}
\begin{lstlisting}[caption=The output of Default from MBPP (problem ID $594$)]
Here is the completed function:

```python
def diff_even_odd(lst):
    first_even = None
    first_odd = None
    
    for num in lst:
        if num % 2 == 0 and first_even is None:
            first_even = num
        elif num % 2 != 0 and first_odd is None:
            first_odd = num
        
        if first_even is not None and first_odd is not None:
            break
    
    if first_even is None or first_odd is None:
        return None
    
    return abs(first_even - first_odd)

# Test case
assert diff_even_odd([1, 3, 5, 7, 4, 1, 6, 8]) == 3
```
\end{lstlisting}
\begin{lstlisting}[caption=The output of CoT from MBPP (problem ID $594$)]
Certainly! Below is a self-contained Python script that defines a function to find the difference between the first even and first odd number of a given list, along with a corresponding test:

```python
def diff_even_odd(numbers):
    first_even = None
    first_odd = None
    
    for number in numbers:
        if number % 2 == 0 and first_even is None:
            first_even = number
        elif number % 2 != 0 and first_odd is None:
            first_odd = number
        
        if first_even is not None and first_odd is not None:
            break
    
    if first_even is None or first_odd is None:
        return None  # Return None if either an even or odd number is not found
    
    return abs(first_even - first_odd)

# Test the function
assert diff_even_odd([1, 3, 5, 7, 4, 1, 6, 8]) == 3
```

This script defines the `diff_even_odd` function, which iterates through the given list to find the first even and first odd numbers. It then calculates the absolute difference between these two numbers and returns it. The test case provided in the problem statement is also included to verify the correctness of the function.
\end{lstlisting}
\begin{lstlisting}[caption=The enriched problem of  MBPP (problem ID $594$)]
Write a function to find the difference of the first even and first odd number of a given list.
assert diff_even_odd([1,3,5,7,4,1,6,8])==3

Analyze the following key terms and their relationships within the problem context:
    [even number]: An integer that is divisible by 2 without leaving a remainder. In the provided list `[1,3,5,7,4,1,6,8]`, the first even number is `4`.
[odd number]: An integer that is not divisible by 2, leaving a remainder of 1. In the provided list `[1,3,5,7,4,1,6,8]`, the first odd number is `1`.
    [difference]: The mathematical operation of subtracting one number from another. In the context of the function, it refers to the result of subtracting the first odd number from the first even number in the given list. For the test case `diff_even_odd([1,3,5,7,4,1,6,8])==3`, the difference is calculated as `4 - 1 = 3`.

assert diff_even_odd([1,3,5,7,4,1,6,8])==3
\end{lstlisting}
\begin{lstlisting}[caption=The output of \SEK from MBPP (problem ID $594$)]
Here is the self-contained Python script that solves the problem described:

```python
def diff_even_odd(numbers):
    """
    Find the difference of the first even and first odd number of a given list.
    
    Args:
    numbers (list): A list of integers.
    
    Returns:
    int: The difference between the first even and first odd number.
    """
    first_even = None
    first_odd = None
    
    for num in numbers:
        if first_even is None and num % 2 == 0:
            first_even = num
        if first_odd is None and num % 2 != 0:
            first_odd = num
        
        if first_even is not None and first_odd is not None:
            break
    
    if first_even is None or first_odd is None:
        raise ValueError("List must contain at least one even and one odd number.")
    
    return first_even - first_odd

# Test case
assert diff_even_odd([1, 3, 5, 7, 4, 1, 6, 8]) == 3
```

This script defines a function `diff_even_odd` that takes a list of integers as input and returns the difference between the first even number and the first odd number in the list. The function iterates through the list to find the first even and first odd numbers and then calculates their difference. The test case provided in the problem statement is also included to verify the function's correctness.
\end{lstlisting}

\section{Details of demonstrations used in KeyExtract \& Explain}

\begin{lstlisting}[caption=The selected demonstrations and corresponding keywords and explanations in Humaneval(+) benchmark]
Demonstration 1:
Check if in given list of numbers, are any two numbers closer to each other than given threshold.
    >>> has_close_elements([1.0, 2.0, 3.0], 0.5)
    False
    >>> has_close_elements([1.0, 2.8, 3.0, 4.0, 5.0, 2.0], 0.3)
    True

[closer to each other]: Describes two numbers in the list whose absolute difference is less than the given threshold. For example, in the list [1.0, 2.8, 3.0, 4.0, 5.0, 2.0] with a threshold of 0.3, the numbers 2.8 and 3.0 are considered closer to each other because |2.8 - 3.0| = 0.2, which is less than 0.3.
[has_close_elements]: Function name that defines the operation to be implemented. It takes two arguments: a list of numbers and a threshold value. The function should return True if any two numbers in the list have a difference smaller than the threshold, and False otherwise.

Demonstration 2:
Input to this function is a string containing multiple groups of nested parentheses. Your goal is to separate those group into separate strings and return the list of those.
    Separate groups are balanced (each open brace is properly closed) and not nested within each other
    Ignore any spaces in the input string.
    >>> separate_paren_groups('( ) (( )) (( )( ))')
    ['()', '(())', '(()())']

[balanced]: Refers to parentheses groups where each opening parenthesis '(' has a corresponding closing parenthesis ')' in the correct order, without any mismatches. Examples of balanced groups include '()', '(())', and '(()())'. In a balanced group, the number of opening and closing parentheses is always equal.
[nested parentheses]: Describes parentheses groups where complete inner pairs are fully contained within outer pairs, without overlapping. The group '(()())' demonstrates this concept, containing two complete inner pairs '()' nested within an outer pair. Nested groups can have multiple levels of nesting while still being balanced.
[separate_paren_groups]: Function name indicating the functionality to be implemented. This function takes a single string argument containing multiple groups of nested parentheses. It should return a list of separated, independent parentheses groups.
\end{lstlisting}
\begin{lstlisting}[caption=The selected demonstrations and corresponding keywords and explanations in MBPP(+)]
Write a function to find the shared elements from the given two lists.
assert set(similar_elements((3, 4, 5, 6),(5, 7, 4, 10))) == set((4, 5))

[shared elements]: Elements that appear in both input lists or sequences. In the test case, 4 and 5 are the shared elements between (3, 4, 5, 6) and (5, 7, 4, 10), as they occur in both sequences.
[similar_elements]: Function name indicating the operation to be implemented. It takes two lists (or tuples) as input and should return a collection of elements common to both input sequences.
\end{lstlisting}
\begin{lstlisting}[caption=The selected demonstrations and corresponding keywords and explanations in APPS]
Demonstration 1:
You have $n$ barrels lined up in a row, numbered from left to right from one. Initially, the $i$-th barrel contains $a_i$ liters of water.

You can pour water from one barrel to another. In one act of pouring, you can choose two different barrels $x$ and $y$ (the $x$-th barrel shouldn't be empty) and pour any possible amount of water from barrel $x$ to barrel $y$ (possibly, all water). You may assume that barrels have infinite capacity, so you can pour any amount of water in each of them. 

Calculate the maximum possible difference between the maximum and the minimum amount of water in the barrels, if you can pour water at most $k$ times.

Some examples:   if you have four barrels, each containing $5$ liters of water, and $k = 1$, you may pour $5$ liters from the second barrel into the fourth, so the amounts of water in the barrels are $[5, 0, 5, 10]$, and the difference between the maximum and the minimum is $10$;  if all barrels are empty, you can't make any operation, so the difference between the maximum and the minimum amount is still $0$. 


-----Input-----

The first line contains one integer $t$ ($1 \le t \le 1000$) — the number of test cases.

The first line of each test case contains two integers $n$ and $k$ ($1 \le k < n \le 2 \cdot 10^5$) — the number of barrels and the number of pourings you can make.

The second line contains $n$ integers $a_1, a_2, \dots, a_n$ ($0 \le a_i \le 10^{9}$), where $a_i$ is the initial amount of water the $i$-th barrel has.

It's guaranteed that the total sum of $n$ over test cases doesn't exceed $2 \cdot 10^5$.


-----Output-----

For each test case, print the maximum possible difference between the maximum and the minimum amount of water in the barrels, if you can pour water at most $k$ times.


-----Example-----
Input
2
4 1
5 5 5 5
3 2
0 0 0

Output
10
0


[barrels]: Containers numbered from 1 to n, where the i-th barrel initially contains a_i liters of water. In the first example, there are 4 barrels, each containing 5 liters of water, represented as [5, 5, 5, 5].
[maximum difference]: The largest possible gap between the fullest and emptiest barrels after performing up to k pourings. For the first example, this value is 10, achieved by creating a barrel with 10 liters and another with 0 liters.

Demonstration 2:
Mikhail walks on a Cartesian plane. He starts at the point $(0, 0)$, and in one move he can go to any of eight adjacent points. For example, if Mikhail is currently at the point $(0, 0)$, he can go to any of the following points in one move:   $(1, 0)$;  $(1, 1)$;  $(0, 1)$;  $(-1, 1)$;  $(-1, 0)$;  $(-1, -1)$;  $(0, -1)$;  $(1, -1)$. 

If Mikhail goes from the point $(x1, y1)$ to the point $(x2, y2)$ in one move, and $x1 \ne x2$ and $y1 \ne y2$, then such a move is called a diagonal move.

Mikhail has $q$ queries. For the $i$-th query Mikhail's target is to go to the point $(n_i, m_i)$ from the point $(0, 0)$ in exactly $k_i$ moves. Among all possible movements he want to choose one with the maximum number of diagonal moves. Your task is to find the maximum number of diagonal moves or find that it is impossible to go from the point $(0, 0)$ to the point $(n_i, m_i)$ in $k_i$ moves.

Note that Mikhail can visit any point any number of times (even the destination point!).


-----Input-----

The first line of the input contains one integer $q$ ($1 \le q \le 10^4$) — the number of queries.

Then $q$ lines follow. The $i$-th of these $q$ lines contains three integers $n_i$, $m_i$ and $k_i$ ($1 \le n_i, m_i, k_i \le 10^{18}$) — $x$-coordinate of the destination point of the query, $y$-coordinate of the destination point of the query and the number of moves in the query, correspondingly.


-----Output-----

Print $q$ integers. The $i$-th integer should be equal to -1 if Mikhail cannot go from the point $(0, 0)$ to the point $(n_i, m_i)$ in exactly $k_i$ moves described above. Otherwise the $i$-th integer should be equal to the the maximum number of diagonal moves among all possible movements.


-----Example-----
Input
3
2 2 3
4 3 7
10 1 9

Output
1
6
-1



-----Note-----

One of the possible answers to the first test case: $(0, 0) \to (1, 0) \to (1, 1) \to (2, 2)$.

One of the possible answers to the second test case: $(0, 0) \to (0, 1) \to (1, 2) \to (0, 3) \to (1, 4) \to (2, 3) \to (3, 2) \to (4, 3)$.

In the third test case Mikhail cannot reach the point $(10, 1)$ in 9 moves.

[revisiting]: The ability to pass through any point, including the destination, multiple times during the journey. In the second example (4, 3, 7), the optimal path includes revisiting coordinates: (0, 0) → (0, 1) → (1, 2) → (0, 3) → (1, 4) → (2, 3) → (3, 2) → (4, 3). This feature allows for maximizing diagonal moves even when the direct path wouldn't utilize all available moves.
\end{lstlisting}

\end{document}